\documentclass[journal]{IEEEtran}
\ifCLASSINFOpdf
\else
\fi

\usepackage{subfigure}
\usepackage{epsfig}
\usepackage{graphicx}
\usepackage{amsmath}
\usepackage{amssymb}
\usepackage{algorithm, algorithmic}
\usepackage{multirow}
\newcommand{\tabincell}[2]{\begin{tabular}{@{}#1@{}}#2\end{tabular}}

\usepackage{url}
\usepackage{color}
\usepackage{amsthm}
\begin{document}
\title{Statistical Loss and Analysis for Deep Learning in Hyperspectral Image Classification}

\author{Zhiqiang~Gong,
        Ping~Zhong,~\IEEEmembership{Senior Member,~IEEE}
        and~Weidong~Hu% <-this % stops a space
\thanks{Manuscript received XX, 2019; revised XX, 2019. This work was supported by the Natural Science Foundation of China under Grant 61671456 and 61971428. (Corresponding author: Ping Zhong)}
\thanks{ Z. Gong is with the National Key Laboratory of Science and Technology on ATR, College of Electrical Science and Technology, National University of Defense Technology, Changsha 410073, China, and also with the National Innovation Institute
of Defense Technology, Chinese Academy of Military Science, Beijing
100000, China. e-mail: (gongzhiqiang13@nudt.edu.cn).}
\thanks{ P. Zhong and W. Hu are with the National Key Laboratory of Science and Technology on ATR, College of Electrical Science and Technology, National University of Defense Technology, Changsha, China, 410073. e-mail: (zhongping@nudt.edu.cn, wdhu@nudt.edu.cn).}% <-this % stops a space
}


\markboth{IEEE LATEX,~Vol X, 2019}%
{Shell \MakeLowercase{\textit{et al.}}: Bare Demo of IEEEtran.cls for IEEE Journals}



\maketitle

\begin{abstract}
 Nowadays, deep learning methods, especially the convolutional neural networks (CNNs), have shown impressive performance on extracting abstract and high-level features from the hyperspectral image. However, general training process of CNNs mainly considers the pixel-wise information or the samples' correlation to formulate the penalization while ignores the statistical properties especially the spectral variability of each class in the hyperspectral image. These samples-based penalizations would lead to the uncertainty of the training process due to the imbalanced and limited number of training samples. To overcome this problem, this work characterizes each class from the hyperspectral image as a statistical distribution and further develops a novel statistical loss with the distributions, not directly with samples for deep learning. Based on the Fisher discrimination criterion, the loss penalizes the sample variance of each class distribution to decrease the intra-class variance of the training samples. Moreover, an additional diversity-promoting condition is added to enlarge the inter-class variance between different class distributions and this could better discriminate samples from different classes in hyperspectral image. Finally, the statistical estimation form of the statistical loss is developed with the training samples through multi-variant statistical analysis.  Experiments over the real-world hyperspectral images show the effectiveness of the developed statistical loss for deep learning.
\end{abstract}

\begin{IEEEkeywords}
Statistical Loss, Deep Learning, Convolutional Neural Networks (CNN), Diversity, Hyperspectral Image classification.
\end{IEEEkeywords}

% \fi
\IEEEpeerreviewmaketitle

\section{Introduction}

With the development of the new and advanced space-borne and aerial-borne sensors, large amounts of hyperspectral images, which contain hundreds of spectral channels, are available \cite{01,02}. The high-dimension spectral bands in the image make it possible to obtain plentiful spectral information to discriminate different objects \cite{52,53,17}. However, { great similarity which occurs in the bands} between different objects makes the image processing task be a challenging one. Besides, the increasing dimensionality in hyperspectral image and the limited number of training samples multiply the difficulties to obtain discriminative features from the image. Therefore, { faced with these circumstances, spatial features are usually incorporated into the representation \cite{04,05}. However,  modelling discriminative spatial and spectral features is not so simple.} There have been increasing efforts to explore effective spectral-spatial methods for hyperspectral image classification.

%Generally, two main strategies are adopted in prior works to incorporate the spatial information in the representation of the hyperspectral image. One is to take the contextual information as additional information to improve the performance by spectral information only. The representative works for this kind of methods are the DBN-CRF model in \cite{16} and the deep pixel-pair features in \cite{15}. The other is to model both the spectral information and the spatial information simultaneously in the training process. Most of the works focus on the research of the latter one since the latter one can make fully use of both the information and adapt the learned model to high-level spatial and spectral information \cite{19,20}. Therefore, this work will mainly develop the spectral-spatial classification method from the aspect of the latter one.

Recently, deep models with multi-layers have demonstrated their potentials in modelling both the spectral and spatial features from the hyperspectral image \cite{23,24,25,16}. Especially, the CNNs, which can capture both the local and the global information from the objects, have presented good performance and been widely applied in hyperspectral image processing tasks. More extended CNNs with the multi-scale convolution \cite{01}, spectral and spatial residual block \cite{20}, have also been developed to improve the representational ability of the CNNs. Therefore, due to the good performance, this work will take advantage of the CNN model to extract the deep spectral-spatial features from the hyperspectral image.

The essential and key problem for the deep representation is how to train a good model. Generally, a good training process is guaranteed by a fine and proper definition of the training loss. The common training loss is constructed with the { training samples directly and  can be broadly divided into two classes.} The first class of losses mainly penalizes the predicted and the real label of each sample for the training of the deep model, such as the generally used softmax loss \cite{15,26}. However, these losses only take advantage of the pixel-wise information from the hyperspectral image while ignore the correlation between different samples.
The other one focuses on the penalization of the samples' correlation \cite{01,22,21}.
These losses penalize the Euclidean distances \cite{27,28} or the angular \cite{51} between sample pairs \cite{30, 31} or among sample triplets \cite{29} and usually provide a better performance than the first one. In real-world applications, the CNN is usually trained under the joint supervisory signals of the losses from the two classes for an effective deep representation.

Even though these samples-based losses have been successfully applied in the training of the deep models, there exist two shortcomings using in the hyperspectral image classification. First, these methods mainly consider the pixel-wise information of each training sample or the pairwise and triplet correlation between different samples which make the training process be susceptible to the imbalanced and limited number of training samples. This would increase the randomness and uncertainty of the training process.  Besides, these methods do not take the statistical properties of the hyperspectral image into consideration. Especially, there exist the spectral variability within each class and the seriously overlapped spectra between different classes in the image. These intrinsic properties could play an important role in providing an effective training process for deep learning.

%Generally, each pixel from the hyperspectral image can be seen as a sampling from each class distribution. Therefore,
To overcome these problems, this work tries to model each class from the image as a certain probabilistic model and formulates the penalization with the class distributions not directly with the samples. The distributions-based loss can reduce the uncertainty caused by the imbalanced and limited number of training samples and further improve the performance of the learned model to extract discriminative features from the image.   Specifically, this work uses the multi-variant normal distributions to model different classes in the image.

Under the probabilistic models and multi-variant statistical analysis, this work develops a novel statistical loss for deep learning in the literature of  hyperspectral image classification.
Based on the Fisher discrimination criterion \cite{50}, the developed statistical loss penalizes the sample variance of each class distribution to decrease the spectral variability of each class. Moreover, a diversity-promoting condition \cite{39} is added in the statistical loss to enlarge the inter-class variance between different class distributions.
Finally, under the multi-variant statistical analysis, the statistical estimation form of the statistical loss is developed with the training samples. As a result, the learned deep model can be more powerful to extract discriminative features from the image.
Overall, the major contributions of this paper are listed as follows.
\begin{itemize}
\item { This work models the hyperspectral image with the probabilistic model and characterizes each class from the image as a certain sampling distribution to take advantage of the statistical properties of the image, so as to formulate the penalization with the class distributions.}
\item Based on the multi-variant statistical analysis and the Fisher discrimination criterion, we develop a novel statistical loss that decreases the spectral variability of each class while enlarges the variance between  different class distributions.
    %Especially, the developed method which treats each class as sampling distribution can be fit for the tasks with limited training samples.
\item Extensive experiments over the real-world hyperspectral image data sets demonstrate the effectiveness and practicability of the developed method and its superiority when compared with other recent samples-based methods.
\end{itemize}

\begin{figure*}[t]
\centering
   \includegraphics[width=0.98\linewidth]{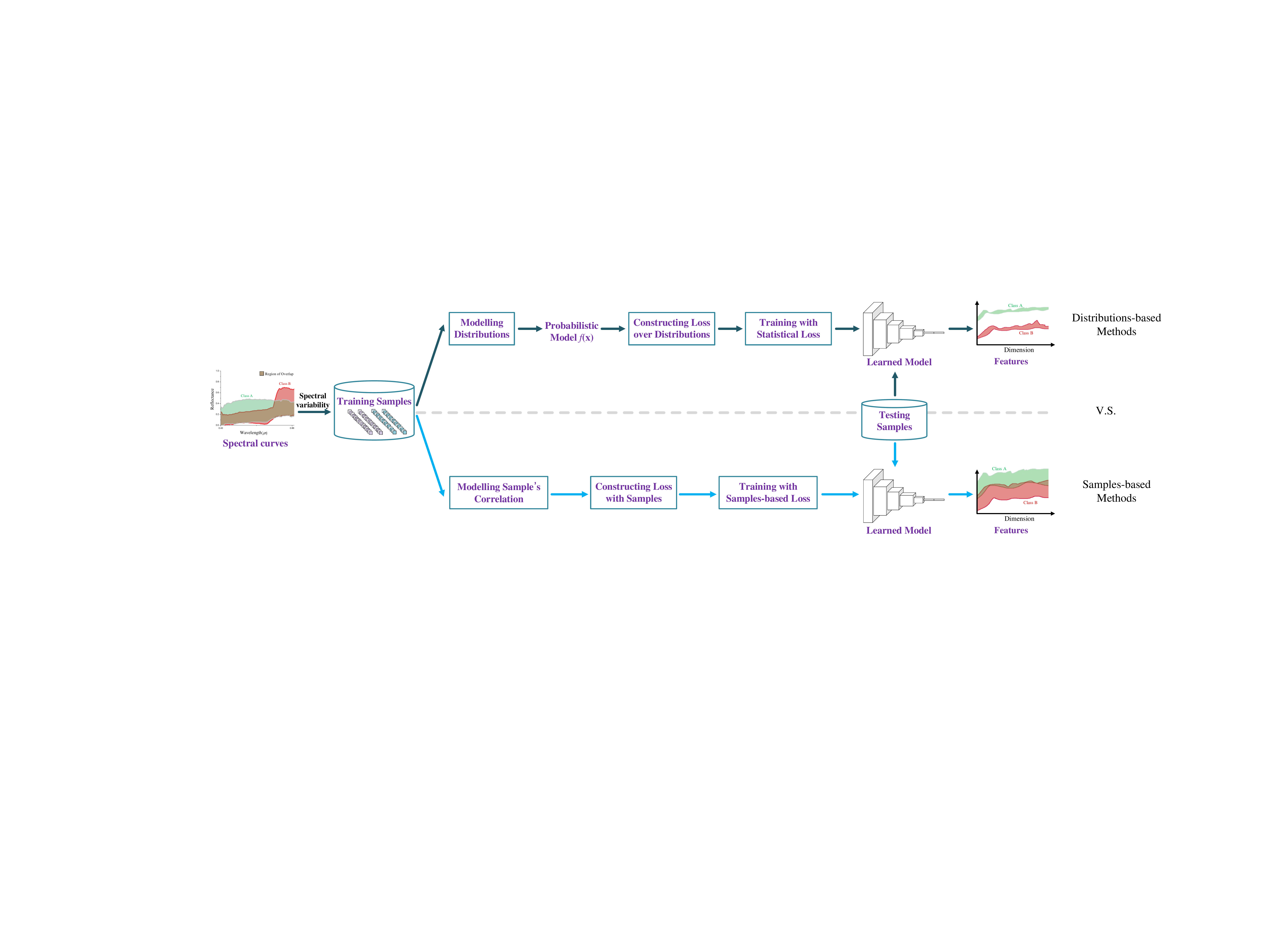}
\vspace{-1ex}
   \caption{Comparison of statistical loss and  samples-based loss. The visualization of spectra curves describes the meadows and bare soil in Pavia University.}
\label{fig:motivation}
\vspace{-2ex}
\end{figure*}

\section{Motivation}

\subsection{Statistical Properties of the Hyperspectral Image} \label{subsec:statistical_property}

Hyperspectral remote sensing measures the radiance of the materials within each pixel area at a very large number of contiguous spectral wavelength bands \cite{38}. The space-borne or aerial-borne sensors gather these spectral information and provide hyperspectral images with hundreds of spectral bands. Since each pixel describes the energy reflected by surface materials and presents the intensity of the energy in different parts of the spectrum, each pixel contains a high-resolution spectrum, which can be used to identify the materials in the pixel by { the analysis} of reflectance or emissivity.

Unfortunately, a theoretically perfect fixed spectrum for any given material does not exist \cite{39}. Due to the variations in the material surface, the spectra observed from samples of the same class are generally not identical. The measured spectra corresponding to pixels with the same class presents an inherent spectral variability that prevents the characterization of homogeneous surface materials by unique spectral signatures. Just as the spectral curves shown in Fig. \ref{fig:motivation}, each class in usual hyperspectral image exhibits remarkable spectral variability and different classes show serious overlapping of the set of spectra.
%Statistics is concerned with making inferences about populations and population characteristics \cite{06,07,08}. Here, population refers to the observations obtained from statistical studies. One of the objectives on statistics is to better understand and model the underlying process which generates the data. Generally, each pixel in the hyperspectral image can be seen as a sampling of each class and obviously, it presents the statistical properties. Therefore, this work tries to describe the features of the hyperspectral image from the statistical view. In the following, we will introduce some essential knowledge of the multivariate statistical analysis and analyze the statistical properties of the hyperspectral image.
%From Fig. \ref{fig:motivation}, we can also find that generally, the tips of vectors of each class corresponding to spectra with variability (known as random spectra) produce a cloud of points.
Besides, most spectra appearing in real applications are random. Therefore, their statistical variability is better described using the probabilistic models $f({\bf x})$.

The learned features from the objects in the image presents the similar characteristics.
{ Since the CNNs have demonstrated their potential in extracting discriminative features from the image \cite{01,09},  this work will use the CNN model} to extract deep features from the image.
The features extracted from the CNNs can be seen as the linear or nonlinear mapping of the objects. Therefore, { the features from the same class also show obvious variability and can be described by the probabilistic models}.

%As defined in \cite{08}, a random variable is a function that associates a real number with each element in the sample space.
%While in real-world applications, random variables are usually related to multiple variants .

For the task at hand, the probabilistic models are { with respect to the high dimensional features}. Therefore, multi-variant statistical analysis, which concerns with analyzing and understanding data in high dimensions, is necessary and just fit { for the image processing task we face with \cite{40}}.
Then, based on the Fisher discrimination criterion and multi-variant statistical analysis, this work will focus on modelling each class from the hyperspectral image as a specific probabilistic model and further develop a novel statistical loss to extract discriminative features from the image.

Even though the hyperspectral image possesses good statistical properties, to the best of our knowledge, this work first takes the statistical properties of the hyperspectral image into consideration and develops the loss with the distributions, not directly with the samples for deep learning. In the following, we will provide a deep comparison between the developed distributions-based loss and the samples-based loss.

\subsection{Distributions-based Loss v.s. Samples-based Loss}

%Deep learning methods have shown potential ability to obtain abstract and high-level features from the objects \cite{44,16,22}.  Many prior works have taken advantage of the CNNs for hyperspectral image, which have obtained discriminative features and presented impressive performance for the task \cite{01,09}.
%As introduced in the former, a good performance of the CNNs is guaranteed by a good and proper training process which can make the learned model be powerful enough to characterize discriminative features from the image.

%Usual CNNs mainly take advantage of the point-to-point information and penalize the real label and the predicted label of each sample via the softmax loss. It usually requires plentiful training samples which can provide enough useful information for a good performance. However, in many real-world applications, especially the hyperspectral image classification which we care about in this work, limited number of training samples are available for the training of the model.
%Under this circumstance, prior works mainly take advantage of the pair-wise correlation between different samples or the triplet information among different samples to further improve the representational ability of the learned CNNs.

The samples-based losses mainly consider the pixel-wise information or penalize the correlation between the sample pairs \cite{27} or triplets \cite{29} for the deep learning. { These losses attempt to obtain good representations of the image} by decreasing the distances between samples from the same class and increasing the distances between samples from different classes.
However, the performance of these samples-based loss is seriously influenced by the imbalanced and limited training samples, which leads to the uncertainty and randomness of the training process.
Fig. \ref{fig:motivation} shows the flowchart of training process by these samples-based loss.  Just as the figure shows, there may exists the overlapping between the obtained features from different classes. Besides, the variability of the learned features from each class would still be too large.

Different from these samples-based losses, the distributions-based loss characterizes each class from the image as a certain probabilistic model and considers the class relationship with the distributions under the Fisher discrimination criterion.
Since we model the correlation based on the class distributions, the problems caused by the imbalance and limitation of the training samples can be solved. This shows positive effects on obtaining discriminative features from the image.
Just as presented in Fig. \ref{fig:motivation}, with the statistical loss by the multi-variant statistical analysis, the spectral variability of the learned features in each class would be decreased and different class distributions can be better separated. This makes the learned features be discriminative enough and thus the classification performance can be significantly improved.
{ In the following, we will introduce the construction of the statistical loss for deep learning in detail.}

%\subsection{Multivariate Statistical Analysis}\label{subsec:distribution}
%
%
%\subsubsection{Multi-variant normal distribution}

%Except for the multi-variant normal distribution, in order to analyze the hyperspectral image from the statistical view, two other famous distributions, namely the $Wishart-distribution$ and Hotelling $T^2-distribution$  are essential. This work will introduce the $Wishart-distribution$ and $T^2-distribution$ in detail based on the requirement of further analysis of hyperspectral image classification.

%\subsubsection{Wishart-distribution}

%\subsubsection{Hotelling $T^2$-distribution}

%\subsection{Statistical Properties of the Hyperspectral Image}

\section{Statistical Loss And Analysis for Deep Learning}

Let us denote $X_0=\{{\bf x}_1,{\bf x}_2,\cdots,{\bf x}_N\}$ as the set of training samples of the hyperspectral image where $N$ is the number of training samples and $y_i$ as the corresponding label of the sample ${\bf x}_i$. $y_i\in Y_0=\{y_{m_1},y_{m_2},\cdots,y_{m_\Lambda}\}$ where $\Lambda$ is the number of the sample classes.

\subsection{Characterizing the Hyperspectral Image with Probabilistic Model}

%From the former section, it can be noted that the hyperspectral image possesses good statistical properties. Moreover, since there exists inherent spectral variability of different samples within each class, it is better to characterize a certain class using probabilistic models.

A reasonable and mostly used probabilistic model for such spectral data in hyperspectral image is generally provided by multivariate normal distribution.
%Since statistical decision procedures, based on normal probability models, are simple and often lead to good performance in modelling target and background, many prior works model the hyperspectral target background spectra \cite{41,43}.
It has already presented impressive performance in modelling target and background as random vectors with multivariate normal distributions for hyperspectral target detection \cite{41} and hyperspectral anomaly detection \cite{43}.
For the task at hand, the extracted features from the CNN model of different classes will also be modelled { with the} multivariate normal distributions.

Given a $p$-dimensional random variable $Z=(Z_1, Z_2, \cdots, Z_p)$ which follows a certain multi-variant distribution.  The random variable ${Z}$ is multi-variant normal if its probability density function (pdf) $f_Z({\bf z})$ has the form
\begin{equation}\label{eq:01}
  f_Z({\bf z})=\frac{1}{(2\pi)^{\frac{p}{2}}|\Sigma|^{\frac{1}{2}}}\exp[-\frac{1}{2}({\bf z}-\mu)^T{\Sigma}^{-1}({\bf z}-\mu)],
\end{equation}
where ${\bf z}=(z_1, z_2, \cdots, z_p)$, $-\infty<z_i<\infty (i=1,2,\cdots,p)$, $\mu$  describes the mean of the distribution and $\Sigma$ which is a positive function represents the covariance matrix of the distribution. Generally, the multi-variant normal distribution can be described as $N_p(\mu,\Sigma)$.

 %The quadratic description of in the exponent of the multivariate normal distribution can be written as \cite{38}
%\begin{equation}\label{eq:05}
%  \Gamma^2=({\bf x}-\mu)^T \Sigma^{-1}({\bf x}-\mu)
%\end{equation}
%It is a statistical distance measure, namely the Mabalanobis distance, which is widely used in the statistical analysis for general hyperspectral image processing task \cite{38,39}.

%

In this work,  each class $k(k=1,2,\cdots,\Lambda)$ is modelled by a certain multi-variant normal distribution with a mean of $\mu_k$ and a covariance of $\Sigma_k$ , which can be written as $N_p(\mu_k,\Sigma_k)$. $p$ represents the dimension of the obtained features from the CNN model and $\Lambda$ denotes the number of classes in the hyperspectral image. Obviously, the sampling distributions corresponding to different classes in the hyperspectral image are independent to each other.

\subsection{Construction of The Statistical Loss} \label{subsec:construction_statistical_loss}

\begin{figure}[t]
\centering
   \includegraphics[width=0.98\linewidth]{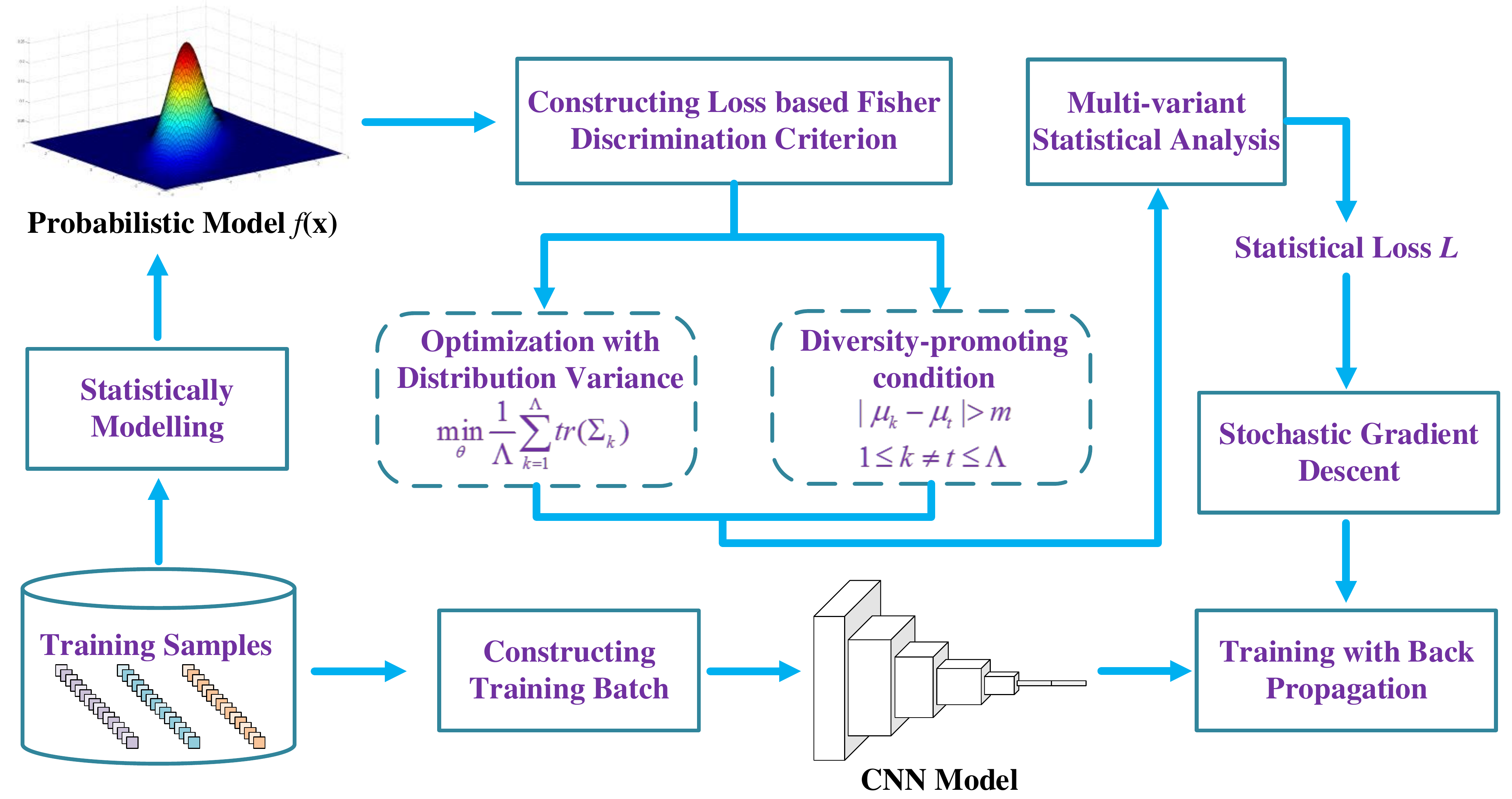}
\vspace{-1ex}
   \caption{Flowchart of the construction of the developed statistical loss.}
\label{fig:statistical_loss}
\vspace{-2ex}
\end{figure}

As Fig. \ref{fig:statistical_loss} shows, this work formulates the loss function based on the Fisher discrimination criterion \cite{50}. Under the criterion, we penalize the sample variance of each class distribution to decrease the intra-class variance, then the problem can be formulated as the following optimization,
\begin{equation}\label{eq:06}
  \min\limits_{\theta}\sum_{k=1}^{\Lambda}tr(\Sigma_k),
\end{equation}
where $tr(\cdot)$ means the trace { of the matrix and $\theta$ denotes} the set of the parameters in the CNN model.

Moreover, to further improve the performance, { this work add additional} diversity-promoting condition to repulse different class distributions from each other. { The diversity-promoting term can be formulated as}
\begin{equation}\label{eq:07}
  |\mu_k-\mu_t|>m,1\leq k\neq t\leq \Lambda,
\end{equation}
where $m$ is a positive value. $k$ and $t$ represents different classes from the image.

Therefore, from the statistical view, we characterize the feature correlation of the hyperspectral image with the probabilistic model and develop the statistical loss as follows,
\begin{equation}\label{eq:08}
\begin{aligned}
  &\min\limits_{\theta}\frac{1}{\Lambda}\sum_{k=1}^{\Lambda}tr(\Sigma_k) \\
  &s.t. |\mu_k-\mu_t|>m,1\leq k\neq t\leq \Lambda.
\end{aligned}
\end{equation}
Under the optimization in Eq. \ref{eq:08}, the intra-class variance of the obtained features is decreased. Besides, the diversity-promoting condition increases the variance between different class distributions. Thus, the learned features can be more discriminative to separate different samples.

To solve the optimization in Eq. \ref{eq:08} with the training samples, { this work statistically estimates the optimization with the multi-variant statistical analysis and develops the estimated statistical loss for hyperspectral image classification.}

\subsection{Statistical Estimation for The Statistical Loss} \label{subsec:statistical_loss}
Generally, in the training process of CNNs, the training batches are usually constructed to accurately estimate the CNN model. A training batch consists of { a batch of randomly selected training samples, which can realize the} parallelization of the training process \cite{22}. Obviously, a training batch can be { looked as a sampling} from the class distributions in the hyperspectral image.

Given a training batch $B$. Denote ${\bf z}_i$ as the feature of ${\bf x}_i$ extracted from the deep model.  $X_k^B=\{{\bf x}_{s_1},{\bf x}_{s_2},\cdots,{\bf x}_{s_{n_k}}\}$ represents the samples of the $k$-th class in the batch. Then, $C_k^B=\{{\bf z}_{s_1},{\bf z}_{s_2},\cdots,{\bf z}_{s_{n_k}}\}$ denotes the extracted features of the $k$-th class where $n_k$ is the number of the samples in the class. Therefore, the features in $C_k^B$ follows the class distribution $N_p(\mu_k,\Sigma_k)$.

\subsubsection{Estimate $\displaystyle{\frac{1}{\Lambda}\sum_{k=1}^{\Lambda}tr(\Sigma_k)}$}
The unbasied estimate $\overline{C}_k$ of the distribution mean $\mu_k$ of the $k$-th class in $B$ can be calculated as
\begin{equation}\label{eq:09}
  \overline{C}_k=\frac{1}{n_k}\sum_{i=1}^{n_k}{\bf z}_{s_i}.
\end{equation}

Define the scatter matrix $S_k$ of the $k$-th class as
\begin{equation}\label{eq:10}
  S_k\triangleq \sum_{j=1}^{n_k}({\bf z}_j-\overline{C}_k)({\bf z}_j-\overline{C}_k)^T.
\end{equation}
Then, for the $k$-th class, the unbiased estimate $\hat{\Sigma}_k$ of the covariance matrix $\Sigma_k$ can be formulated as
\begin{equation}\label{eq:11}
  \hat{\Sigma}_k=\frac{1}{n_k-1}S_k=\frac{1}{n_k-1}\sum_{j=1}^{n_k}({\bf z}_j-\overline{C}_k)({\bf z}_j-\overline{C}_k)^T.
\end{equation}
{ We use $\hat{\Sigma}_k$ to estimate the covariance matrix $\Sigma_k$. Then,
\begin{equation}\label{eq:12}
  tr(\hat{\Sigma}_k)=\frac{1}{n_k-1}\sum_{j=1}^{n_k}({\bf z}_j-\overline{C}_k)^T({\bf z}_j-\overline{C}_k).
\end{equation}
 Besides, $\displaystyle{\frac{1}{\Lambda}\sum_{k=1}^{\Lambda}tr(\Sigma_k)}$ can be estimated by $\displaystyle{\frac{1}{\Lambda}\sum_{k=1}^{\Lambda}tr(\hat{\Sigma}_k)}$. Therefore, $\displaystyle{\frac{1}{\Lambda}\sum_{k=1}^{\Lambda}tr(\Sigma_k)}$ is estimated by}
\begin{equation}\label{eq:13}
  {\frac{1}{\Lambda}\sum_{k=1}^{\Lambda}tr(\hat{\Sigma}_k)}=\frac{1}{\Lambda}\sum_{k=1}^{\Lambda}[\frac{1}{n_k-1}\sum_{j=1}^{n_k}({\bf z}_j-\overline{C}_k)^T({\bf z}_j-\overline{C}_k)].
\end{equation}

\subsubsection{Estimate $|\mu_k-\mu_t|>m$}
Given the $k$-th and the $t$-th class. The $k$-th class follows the multi-variant normal distribution as $C^B_k\sim N_p(\mu_k,\Sigma_k)$ and the $t$-th class follows $C^B_t\sim N_p(\mu_t,\Sigma_t)$. Obviously, the two class distributions are independent from each other. This work will use the statistical hypothesis to estimate the condition $|\mu_k-\mu_t|>m$.

To { estimate $|\mu_k-\mu_t|>m$}, two famous multi-variant distributions, namely the $Wishart-distribution$ and the Hotelling $T^2-distribution$ are necessary.

The $Wishart-distribution$ plays a prominent role in the analysis of estimated covariance matrices. Assume $u_1,u_2,\cdots,u_n$ as independent distributions which follows the same $p$-dimensional multi-variant normal distribution $N_p(0,\Sigma)$. Denote $u=(u_1,u_2,\cdots,u_n)$. Then the random matrix $W=uu^T=\sum_{i=1}^{n}u_iu_i^T$ follows the $p$-dimensional $Wishart-distribution$ with $n$ degrees of freedom, which can be written as
\begin{equation}\label{eq:02}
  W=\sum_{i=1}^{n}u_iu_i^T\sim W_p(n,\Sigma).
\end{equation}
It should be noted that the $Wishart-distribution$ satisfies the following property. If statistics $W_i\sim W_p(n_i,\Sigma),i=1,2,\cdots,k$, and the statistics $W_i(i=1,2,\cdots,k)$ are independent from each other, then,
\begin{equation}\label{eq:03}
  W=\sum_{i=1}^{k}W_i\sim W_p(\sum_{i=1}^{k}n_i,\Sigma).
\end{equation}

The Hotelling $T^2-distribution$ { is essential to the hypothesis testing} in multi-variant statistical analysis. Suppose that $X\sim N_p(\mu,\Sigma)$ is { independent to $W\sim W_p(n,\Sigma)$. Denote the} statistic $T^2=n(X-\mu)^TW^{-1}(X-\mu)$, then the statistic $T^2$ is defined as the Hotelling $T^2-distribution$ with $n$ degree of freedom, which can be formulated as
\begin{equation}\label{eq:04}
  T^2=n(X-\mu)^TW^{-1}(X-\mu)\sim T^2(p,n).
\end{equation}
It should be noted that the former $Wishart-distribution$ and Hotelling $T^2-distribution$ are certain { distributions where the probability distribution is fixed under a certain degrees of freedom}. In the following, the two distributions will play an important role in { the following estimation}.

Traditionally, a statistical hypothesis is an assertion or conjecture concerning one or more populations. It should be noted that the rejection of a hypothesis implies that the sample evidence refutes it. That is to say, rejection means that there is a small probability of obtaining the sample information observed when, in fact, the hypothesis is true \cite{06}. The structure of hypothesis testing will be formulated with the use of the null hypothesis $H_0$ and the alternative hypothesis $H_1$. Generally, the rejection of $H_0$ leads to the acceptance of the alternative hypothesis $H_1$.

For simplicity, this work would set the $m$ in Eq. \ref{eq:08} to 0.
Therefore, from the statistical hypothesis view, we may then re-state the condition $|\mu_k-\mu_t|>m$ as the following two competing hypotheses:
\begin{align}\label{eq:14_15}
  H_0: \mu_k=\mu_t, \\
  H_1: \mu_k\neq\mu_t.
\end{align}
The scatter matrices $S_k$ and $S_t$ of the $k$-th and the $t$-th class are defined as Eq. \ref{eq:10} shows. As the definition of $Wishart-distribution$, it can be noted that
\begin{align}\label{eq:16_17}
  S_k\sim W_p(n_k-1,\Sigma_k), \\
  S_t\sim W_p(n_t-1,\Sigma_t).
\end{align}
Since all the samples of different classes are from the same hyperspectral image, just as processed in many hyperspectral target recognition task \cite{38}, different class distributions are supposed to have the same covariance matrix, { namely} $\Sigma_k=\Sigma_t=\Sigma$. Therefore, based on the properties of $Wishart-distribution$ as Eq. \ref{eq:03}, the statistic $S_k+S_t$ follows
\begin{equation}\label{eq:18}
  S_k+S_t\sim W_p(n_k+n_t-2,\Sigma).
\end{equation}
Moreover, { depending on} the definition of the multi-variant normal distribution, we can find that the statistic $M_{k,t}$ which is defined as $M_{k,t}=\overline{C}_k-\overline{C}_t-(\mu_k-\mu_t)$ follows the multi-variant normal distribution,
\begin{equation}\label{eq:19}
  M_{k,t}=\overline{C}_k-\overline{C}_t-(\mu_k-\mu_t)\sim N_p(0,(\frac{1}{n_k}+\frac{1}{n_t})\Sigma).
\end{equation}
Furthermore, denote the statistic $T^2$
\begin{equation}\label{eq:20}
  T^2=\frac{n_k+n_t-2}{\frac{1}{n_k}+\frac{1}{n_t}}M_{k,t}^T(S_k+S_t)^{-1}M_{k,t}.
\end{equation}
Then, according to { the definition of Hotelling $T^2-distribution$, it can be noted that} the statistic $T^2$ in Eq. \ref{eq:20} follows the $T^2-distribution$ as
\begin{equation}\label{eq:21}
  T^2\sim T^2(p,n_k+n_t-2).
\end{equation}
Therefore, at the $\alpha$ level of confidence, if $T^2\leq T^2_{p,n_k+n_t-2}(\alpha)$, accept the null hypothesis $H_0$, reject the alternative hypothesis $H_1$; otherwise, if $T^2> T^2_{p,n_k+n_t-2}(\alpha)$, accept the alternative hypothesis $H_1$, reject the null hypothesis $H_0$.

Since the alternative hypothesis $H_1$ is what we seek, { $|\mu_k-\mu_t|>m$ can be transformed} to the following one,
\begin{equation}\label{eq:22}
  \frac{n_k+n_t-2}{\frac{1}{n_k}+\frac{1}{n_t}}(\overline{C}_k-\overline{C}_t)^T(S_k+S_t)^{-1}(\overline{C}_k-\overline{C}_t)>T^2_{p,n_k+n_t-2}(\alpha).
\end{equation}

\subsubsection{Formulate the Statistical Loss}
Denote $\Gamma_{k,t}=\overline{C}_k-\overline{C}_t$. Then, based on the Eq. \ref{eq:13} and Eq. \ref{eq:22}, the optimization problem in Eq. \ref{eq:08} can be transformed as
\begin{equation}\label{eq:23}
  \begin{aligned}
  &\min\frac{1}{\Lambda}\sum_{k=1}^{\Lambda}(\frac{1}{n_k-1}\sum_{j=1}^{n_k}({\bf z}_j-\overline{C}_k)^T({\bf z}_j-\overline{C}_k)) \\
  &s.t.\frac{n_k+n_t-2}{\frac{1}{n_k}+\frac{1}{n_t}}\Gamma_{k,t}^T(S_k+S_t)^{-1}\Gamma_{k,t}>T^2_{p,n_k+n_t-2}(\alpha), \\
  &\ \ \ \  1\leq k\neq t\leq \Lambda.
  \end{aligned}
\end{equation}
By Lagrange multiplier, the statistical loss for the hyperspectral image can be formulated from Eq. \ref{eq:23} as
\begin{equation}\label{eq:24}
  \begin{aligned}
  L=&\frac{1}{\Lambda}\sum_{k=1}^{\Lambda}(\frac{1}{n_k-1}\sum_{j=1}^{n_k}({\bf z}_j-\overline{C}_k)^T({\bf z}_j-\overline{C}_k))+ \\
  &\lambda\sum_{k\neq t}^{\Lambda}(T^2_{p,n_k+n_t-2}(\alpha)-\frac{n_k+n_t-2}{\frac{1}{n_k}+\frac{1}{n_t}}\Gamma_{k,t}^T(S_k+S_t)^{-1}\Gamma_{k,t}),
  \end{aligned}
\end{equation}
where $\lambda$ is the tradeoff parameter.

Besides, $T^2_{p,n_k+n_t-2}(\alpha)$ is a constant value that is irrelevant to the training samples, therefore, we set $T^2_{p,n_k+n_t-2}(\alpha)$ as a constant positive value. Then, Eq. \ref{eq:24} can be re-formulated as
\begin{equation}\label{eq:25}
  \begin{aligned}
  L=&\frac{1}{\Lambda}\sum_{k=1}^{\Lambda}(\frac{1}{n_k-1}\sum_{j=1}^{n_k}({\bf z}_j-\overline{C}_k)^T({\bf z}_j-\overline{C}_k))+ \\
  &\lambda\sum_{k\neq t}^{\Lambda}(\Delta-\frac{n_k+n_t-2}{\frac{1}{n_k}+\frac{1}{n_t}}\Gamma_{k,t}^T(S_k+S_t)^{-1}\Gamma_{k,t}),
  \end{aligned}
\end{equation}
where $\Delta$ represents a positive value. Therefore, Eq. \ref{eq:25} defines the { statistical loss for deep learning in this work}. Fig. \ref{fig:statistical_loss} shows the detailed process to formulate the statistical loss. Under the statistical loss, the learned model can  be more discriminative for the hyperspectral image.

%Fig. \ref{fig:statistical_loss} shows the flowchart of the construction of the statistical loss for hyperspectral image. The statistical loss calculates the inter-class and intra-class variance between the samples in the training batch and is easy to implement. Moreover, the statistical loss measures the difference from the class view which can solve the unbalanced training from unbalanced data as well as the task with limited training samples.

\section{Training}
Generally, the deep model is trained with the stochastic gradient descent method and back propagation is used for the training process of the model \cite{10}.  Therefore, the main problem for the implementation of the developed statistical loss { in hyperspectral image classification task} is to compute the { derivation} of the statistical loss w.r.t. the extracted features from the training samples.

As defined in section \ref{subsec:statistical_loss}, the statistical loss can be formulated as
\begin{equation}\label{eq:26}
  L=L_0+\lambda L_{div},
\end{equation}
where
\begin{align}
 \label{eq:27} L_0&=\frac{1}{\Lambda}\sum_{k=1}^{\Lambda}(\frac{1}{n_k-1}\sum_{j=1}^{n_k}({\bf z}_j-\overline{C}_k)^T({\bf z}_j-\overline{C}_k)), \\
 \label{eq:28} L_{div}&=\sum_{k\neq t}^{\Lambda}(\Delta-\frac{n_k+n_t-2}{\frac{1}{n_k}+\frac{1}{n_t}}\Gamma_{k,t}^T(S_k+S_t)^{-1}\Gamma_{k,t}).
\end{align}
According to the chain rule, gradients of the statistical loss w.r.t. ${\bf z}_i$ can be formulated as
\begin{equation}\label{eq:gradients_statistical_loss}
  \frac{\partial L}{\partial {\bf z}_i}=\frac{\partial L_0}{\partial {\bf z}_i}+\lambda\frac{\partial L_{div}}{\partial {\bf z}_i}.
\end{equation}

{ The partial of $L_0$ w.r.t. ${\bf z}_i$ can be easily computed by}
\begin{equation}\label{eq:29}
  \frac{\partial L_0}{\partial {\bf z}_i}=\frac{2}{\Lambda}\sum_{k=1}^{\Lambda}\frac{1}{n_k}I({\bf z}_i\in C_k^B)({\bf z}_i-\overline{C}_k),
\end{equation}
where ${\bf z}_i$ is the learned features of training sample ${\bf x}_i$ from the CNN model. $I(\cdot)$ denotes the indicative function.

Besides, the partial of $L_{div}$ w.r.t. ${\bf z}_i$ can be calculated by
\begin{equation}\label{eq:30}
  \frac{\partial L_{div}}{\partial {\bf z}_i}=-\sum_{k\neq t}^{\Lambda}\frac{n_k+n_t-2}{\frac{1}{n_k}+\frac{1}{n_t}}\frac{\partial \Gamma_{k,t}^T(S_k+S_t)^{-1}\Gamma_{k,t}}{\partial {\bf z}_i}.
\end{equation}
Therefore, the { key process} is to calculate the following derivation:
\begin{equation}\label{eq:31}
\begin{aligned}
  \frac{\partial \Gamma_{k,t}^T(S_k+S_t)^{-1}\Gamma_{k,t}}{\partial {\bf z}_i}=&\frac{\partial \Gamma_{k,t}^T(S_k+S_t)^{-1}}{\partial {\bf z}_i}\Gamma_{k,t} \\
  &+\frac{\partial \Gamma_{k,t}^T}{\partial {\bf z}_i}(S_k+S_t)^{-1}\Gamma_{k,t}.
\end{aligned}
\end{equation}
The $\displaystyle{\frac{\partial \Gamma_{k,t}^T}{\partial {\bf z}_i}}$ { can be computed as}
\begin{equation}\label{eq:32}
  \frac{\partial \Gamma_{k,t}^T}{\partial {\bf z}_i}=\frac{\partial (\overline{C}_k-\overline{C}_t)^T}{\partial {\bf z}_i}=\frac{1}{n_k}I({\bf z}_i\in C_k)I_0,
\end{equation}
where $I_0$ represents the identity matrix. In addition,
\begin{equation}\label{eq:33}
\begin{aligned}
  &\frac{\partial [\Gamma_{k,t}^T(S_k+S_t)^{-1}]}{\partial {\bf z}_i}=\frac{\partial \Gamma_{k,t}^T}{\partial {\bf z}_i}(S_k+S_t)^{-1} \\
  &-\frac{n_k-1}{n_k}I({\bf z}_i\in C_k^B)[(\overline{C}_k-\overline{C}_t)^T(S_k+S_t)^{-1}{\bf z}_i](S_k+S_t)^{-1} \\
  &-\frac{n_k-1}{n_k}I({\bf z}_i\in C_k^B)(S_k+S_t)^{-1}(\overline{C}_k-\overline{C}_t){\bf z}_i^T(S_k+S_t)^{-1}.
  \end{aligned}
\end{equation}

{ Based on Eqs.} \ref{eq:30}, \ref{eq:31}, \ref{eq:32} and \ref{eq:33}, the partial of $L_{div}$ w.r.t. ${\bf z}_i$ can be calculated by\footnote{Detailed computations of gradients are shown in the supplemental materials.}
\begin{equation}\label{eq:34}
  \begin{aligned}
&\frac{\partial L_{div}}{\partial {\bf z}_i}=\frac{n_k+n_t-2}{\frac{1}{n_k}+\frac{1}{n_t}}[-\frac{2}{n_k}I({\bf z}_i\in C_k^B)(S_k+S_t)^{-1}(\overline{C}_k-\overline{C}_t)\\
  &+\frac{n_k-1}{n_k}I({\bf z}_i\in C_k^B)[(\overline{C}_k-\overline{C}_t)^T(S_k+S_t)^{-1}{\bf z}_i](S_k+S_t)^{-1}\\
  &+\frac{n_k-1}{n_k}I({\bf z}_i\in C_k^B)(S_k+S_t)^{-1}(\overline{C}_k-\overline{C}_t){\bf z}_i^T(S_k+S_t)^{-1}].
  \end{aligned}
\end{equation}
Through back propagation with the former equations, the CNN model can be trained with the training samples and discriminative features can be learned from the hyperspectral image. The detailed training process of the developed method is shown in Algorithm \ref{algorithm:1}. It should also be noted that the whole CNN is trained under joint supervisory signals of softmax loss and our statistical loss.

\begin{algorithm}[t]
\renewcommand{\algorithmicrequire}{\textbf{Input:}}
\renewcommand{\algorithmicensure}{\textbf{Output:}}
\caption{Training process of the developed method}\label{algorithm:1}
\begin{algorithmic}[1]
\REQUIRE ${\bf x}_i (i=1,2,\cdots,N)$, $\lambda,\beta,\Delta$, $\theta_k=\{W_k,{\bf b}_k\}$ as the parameters of $k$-th layer, $\theta_0=\{W_0,{\bf b}_0\}$ as the parameters in Softmax layer, learning rate $lr$.
\ENSURE $\theta_k$, $W_0, {\bf b}_0$
\STATE Initialize $\theta_k$ in $k$-th convolutional layer where $W_k$ is initialized from Gaussian distribution with standard derivation of 0.01 and ${\bf b}_k$ is set to 0.
\WHILE{not converge}
\STATE $t \leftarrow t+1$.
\STATE Construct the training batch $B^t$ randomly.
\STATE Obtain the deep features ${\bf z}_i^t$ from the sample ${\bf x}_i^t\in B^t$ with the CNN model specified by $\theta_k^t$.
\STATE Compute the penalization of $L^t_0$ using Eq. \ref{eq:27}.
\STATE Compute the penalization of the diversity-promoting term $L^t_{div}$ using Eq. \ref{eq:28}.
\STATE Compute the statistical loss by $L^t=L^t_0+\lambda L^t_{div}$.
\STATE Compute the joint loss by $L^t_{joint}=L^t_s+\beta L^t$ where $L^t_s$ is the penalization from the softmax loss and $\beta$ is the tradeoff parameter.
\STATE Compute the { derivation} of $L^t_0$ w.r.t. ${\bf z}_i^t$ in $B^t$ using Eq. \ref{eq:29}.
\STATE Compute the { derivation} of $L^t_{div}$ w.r.t. ${\bf z}_i^t$ in $B^t$ as Eq. \ref{eq:34} shows.
\STATE Update the parameters $\theta_0$ by \\ ${\theta_0^{t+1}=\theta_0^{t}-lr\times \frac{\partial L^t_{joint}}{\partial \theta_0^{t}}=}$ ${\theta_0^{t}-lr\times \frac{\partial L^t_{s}}{\partial \theta_0^{t}}}$.
\STATE Update the parameters $\theta_k$ by \\ $\theta_k^{t+1}=\theta_k^{t}-lr\times \frac{\partial L_{joint}^t}{\partial \theta_k^t}=\theta_k^{t}-lr\times \frac{\partial L_{joint}^t}{\partial {\bf z}_i^t}\times\frac{\partial {\bf z}_i^t}{\partial \theta_k^t}$.
\ENDWHILE
\RETURN $\theta_k$, $\theta_0=\{W_0, {\bf b}_0\}$
\end{algorithmic}
\end{algorithm}

\section{Experimental Results}

\subsection{Experimental Datasets and Experimental Setups}\label{subsec:setup}
To further validate the effectiveness of the developed statistical loss, this work conducts experiments over the real-world hyperspectral image data sets, namely Pavia University and Indian Pines\footnote{More results can be seen in the supplemental materials.}. We also compare the experimental results with other state-of-the-art methods including the most recent samples-based loss to show the advantage of the proposed method. In addition, overall average (OA), average accuracy (AA), and Kappa are chosen as the { measurements} to evaluate the classification performance. All the results in this work come from the average value and standard deviation of ten runs of training and testing. { For each of the ten experiments, the training and testing sets are randomly selected.}

Pavia University data \cite{11} was gathered by the reflective optics system imaging spectrometer (ROSIS-3) sensor with a spatial resolution of 1.3m per pixel. It consists of $610\times 340$ pixels of which a total of 42, 776 labelled samples divided into nine classes have been chosen for experiments. Each pixel denotes a sample and consists of 115 bands with a spectral coverage ranging from 0.43 to 0.86 $\mu m$. 12 spectral bands are abandoned due to the noise and the remaining 103 channels are used for experiments. %The false color and groundtruth of the data are shown in Fig. \ref{fig:pavia}.

%\begin{figure}[t]
%\centering
% \subfigure[]{\includegraphics[width=0.37\linewidth]{pavia_1.jpg}}
% \subfigure[]{\includegraphics[width=0.37\linewidth]{data/pavia_groundtruth.jpg}}
% \subfigure[]{\includegraphics[width=0.23\linewidth]{data/pavia_groundtruth.pdf}}
%\vspace{-3ex}
%   \caption{Pavia University data. (a) False color composite (band 10, 60, 90); (b) ground truth; (c) map color.}
%\label{fig:pavia}
%\vspace{-2ex}
%\end{figure}
%
%\begin{figure}[t]
%\centering
% \subfigure[]{\includegraphics[width=0.355\linewidth]{data/indian_1.jpg}}
% \subfigure[]{\includegraphics[width=0.355\linewidth]{data/indian_groundtruth.jpg}}
% \subfigure[]{\includegraphics[width=0.255\linewidth]{data/indian_groundtruth.pdf}}
%\vspace{-3ex}
%   \caption{Indian Pines data. (a) False color composite (band 30, 60, 90); (b) ground truth; (c) map color.}
%\label{fig:indian}
%\vspace{-4ex}
%\end{figure}

Indian Pines data \cite{12} was collected by the 224-band AVIRIS sensor ranging from 0.4 to 2.5 $\mu m$ over the Indian Pines test site in north-western Indiana. It consists of $145\times 145$ pixels and the corrected data of Indian Pines remains 200 bands where 24 bands covering the region of water absorption are removed. Sixteen land-cover classes with a total of 10249 labelled samples are selected from the data for experiments.
%(See Fig. \ref{fig:indian} for details)

\begin{figure*}[t]
\centering
   \includegraphics[width=0.92\linewidth]{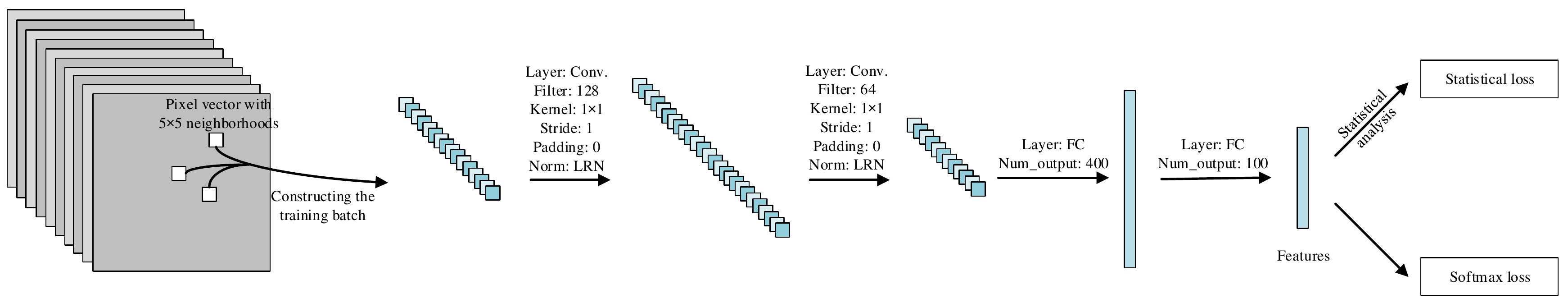}
\vspace{-1ex}
   \caption{The deep structure  adopted in this work to implement the proposed method for hyperspectral image Classification. The whole CNN is trained under the joint supervisory signals of softmax loss and our statistical loss.}
\label{fig:cnn_model}
\vspace{-2ex}
\end{figure*}

Caffe is chosen as the deep learning framework to implement the proposed method \cite{34}. Since this work mainly test the effectiveness of the developed statistical loss, we will use the CNN model just as Fig. \ref{fig:cnn_model} shows for all the experiments in this work. The learning rate, epoch iteration, training batch are set to 0.001, 60000, 84, respectively. As Fig. \ref{fig:cnn_model} shows, this work uses the $5\times 5$ neighborhoods to incorporate the spatial information. In the experiments, we choose 200 samples per class for training and the remainder for testing over Pavia University while over Indian Pines data, we select 20\% of samples per class for training.
{ The code for the implementation of the proposed method will be released soon at \url{http://github.com/shendu-sw/statistical-loss}}.

\subsection{General Performance}
At first, we present a brief overview of the merits of the developed statistical loss for hyperspectral image classification. In this set of experiments, the diversity weight $\lambda$ is fixed as constant 0.01. { General machine with a 4.00GHz Intel $\circledR$ Core (IM) i7-6700K CPU, 64 GB memory, and GeForce GTX 1080 GPU is chosen to perform the proposed method. The proposed method implemented through caffe took about 1146s over Pavia University and 1610s over Indian Pines data. It should be noted that this work implements the developed statistical loss by CPU and the computational performance can be remarkably improved by modifying the codes to run the developed method on GPUs}.

\begin{table}[t]
\begin{center}
\caption{Classification accuracies ($Mean\pm SD$) (OA, AA, and Kappa) of different methods achieved on the Pavia University data. The results from CNN is trained with the Softmax Loss. ${|F_{ij}|}$ represents the value of McNemar's test.}
\label{table:pavia}
\vspace{-1ex}
\begin{tabular}{|c | c | c c c|}
%\toprule[1pt]
\hline
\multicolumn{2}{|c|}{\bf Methods}     &  {\bf SVM-POLY} &  {\bf CNN} &  {\bf Proposed Method} \\
\hline\hline
\multirow{9}{*}{\rotatebox{90}{\tabincell{c}{\textbf{Classification} \\ \textbf{Accuracies (\%)}}}}          & C1   &  $83.01\pm 1.30$  &  $98.50\pm 0.49$ &  $\mathbf{99.59\pm 0.30}$\\
                                                                                                             & C2   &  $86.61\pm 1.80$  &  $99.02\pm 0.59$ &  $\mathbf{99.72\pm 0.11}$\\
                                                                                                             & C3   &  $85.96\pm 1.04$  &  $95.92\pm 2.84$ &  $\mathbf{96.84\pm 1.38}$\\
                                                                                                             & C4   &  $96.36\pm 0.92$  &  $98.78\pm 0.55$ &  $\mathbf{99.39\pm 0.48}$\\
                                                                                                             & C5   &  $99.62\pm 0.18$  &  $100.0\pm 0.00$ &  $\mathbf{100.0\pm 0.00}$\\
                                                                                                             & C6   &  $90.96\pm 1.57$  &  $99.36\pm 1.00$ &  $\mathbf{99.70\pm 0.33}$\\
                                                                                                             & C7   &  $93.92\pm 0.80$  &  $99.56\pm 0.36$ &  $\mathbf{99.96\pm 0.06}$\\
                                                                                                             & C8   &  $87.27\pm 1.56$  &  $95.90\pm 3.31$ &  $\mathbf{99.13\pm 0.62}$\\
                                                                                                             & C9   &  $99.93\pm 0.13$  &  $100.0\pm 0.00$ &  $\mathbf{100.0\pm 0.00}$\\
 \hline
 \multicolumn{2}{|c|}{{\bf OA}  (\%)}      &  $88.07\pm 0.82$ &  $98.61\pm 0.35$  &  $\mathbf{99.51\pm 0.09}$\\
 \hline
 \multicolumn{2}{|c|}{{\bf AA}  (\%)}      &  $91.52\pm 0.26$ &  $98.56\pm 0.36$  &  $\mathbf{99.37\pm 0.13}$\\
 \hline
 \multicolumn{2}{|c|}{{\bf KAPPA} (\%)}    &  $84.35\pm 1.01$ &  $98.14\pm 0.47$  &  $\mathbf{99.34\pm 0.12}$\\
\hline
\multicolumn{2}{|c|}{{${|F_{ij}|}$}}    &  $49.28$ &  $15.50$  &  $-$\\
\hline
%\bottomrule[1pt]
\end{tabular}
\end{center}
\vspace{-2ex}
\end{table}

\begin{table}[t]
\begin{center}
\caption{Classification accuracies (OA, AA, and Kappa) of different methods achieved on the Indian Pines data.}
\label{table:indian}
\vspace{-1ex}
\begin{tabular}{| c | c | c c c |}
%\toprule[1pt]
\hline
\multicolumn{2}{|c|}{\bf Methods}     &  {\bf SVM-POLY} &  {\bf CNN} &  {\bf Proposed Method} \\
\hline\hline
\multirow{9}{*}{\rotatebox{90}{\tabincell{c}{\textbf{Classification} \\ \textbf{Accuracies (\%)}}}}          & C1   &  $82.78\pm 6.11$  &  $96.11\pm 3.26$ &  $\mathbf{96.11\pm 2.68}$\\
                                                                                                             & C2   &  $82.65\pm 1.89$  &  $99.27\pm 0.37$ &  $\mathbf{99.43\pm 0.41}$\\
                                                                                                             & C3   &  $77.15\pm 2.22$  &  $98.86\pm 1.13$ &  $\mathbf{99.68\pm 0.34}$\\
                                                                                                             & C4   &  $74.29\pm 5.65$  &  $\mathbf{98.68\pm 1.50}$ &  ${97.72\pm 2.80}$\\
                                                                                                             & C5   &  $91.79\pm 2.15$  &  $98.47\pm 1.12$ &  $\mathbf{99.53\pm 0.56}$\\
                                                                                                             & C6   &  $97.50\pm 1.31$  &  $99.69\pm 0.28$ &  $\mathbf{100.0\pm 0.00}$\\
                                                                                                             & C7   &  $85.45\pm 7.67$  &  $\mathbf{99.09\pm 1.92}$ &  ${95.91\pm 7.56}$\\
                                                                                                             & C8   &  $99.63\pm 0.31$  &  $99.87\pm 0.41$ &  $\mathbf{100.0\pm 0.00}$\\
                                                                                                             & C9   &  $55.00\pm 12.4$  &  $\mathbf{99.38\pm 1.98}$ &  ${91.88\pm 8.86}$\\
                                                                                                             & C10   &  $84.52\pm 1.60$  &  $98.69\pm 0.75$ &  $\mathbf{99.60\pm 0.53}$\\
                                                                                                             & C11   &  $90.73\pm 0.78$  &  $99.04\pm 0.51$ &  $\mathbf{99.61\pm 0.35}$\\
                                                                                                             & C12   &  $88.25\pm 2.55$  &  $98.78\pm 0.84$ &  $\mathbf{99.43\pm 0.32}$\\
                                                                                                             & C13   &  $97.99\pm 2.05$  &  $\mathbf{99.57\pm 0.58}$ &  ${99.21\pm 0.50}$\\
                                                                                                             & C14   &  $96.50\pm 0.56$  &  $99.66\pm 0.33$ &  $\mathbf{99.77\pm 0.21}$\\
                                                                                                             & C15   &  $67.66\pm 3.75$  &  $96.07\pm 2.95$ &  $\mathbf{98.47\pm 1.75}$\\
                                                                                                             & C16   &  $88.92\pm 6.07$  &  $\mathbf{99.32\pm 0.96}$ &  ${98.11\pm 3.20}$\\
 \hline
 \multicolumn{2}{|c|}{{\bf OA}  (\%)}      &  $88.20\pm 0.51$ &  $99.03\pm 0.28$  &  $\mathbf{99.49\pm 0.13}$\\
 \hline
 \multicolumn{2}{|c|}{{\bf AA}  (\%)}      &  $85.05\pm 1.26$ &  $\mathbf{98.79\pm 0.45}$  &  ${98.40\pm 0.93}$\\
 \hline
 \multicolumn{2}{|c|}{{\bf KAPPA} (\%)}    &  $86.49\pm 0.58$ &  $98.89\pm 0.32$  &  $\mathbf{99.42\pm 0.15}$\\
\hline
\multicolumn{2}{|c|}{{${|F_{ij}|}$}}    &  $30.10$ &  $4.48$  &  $-$\\
\hline
%\bottomrule[1pt]
\end{tabular}
\end{center}
\vspace{-2ex}
\end{table}

Tables \ref{table:pavia} and \ref{table:indian} show the classification results over the two datasets separately.
{ For Pavia University data, C1, C2, $\cdots$, C9 represent the asphalt, meadows, gravel, trees, metal sheet, bare soil, bitumen, brick, shadow, respectively. For Indian Pines data, C1, C2, $\cdots$, C16 stand for the alfalfa, corn-no-till, corn-min-till, corn, grass\_pasture, grass\_trees, grass\_pasture-mowed, hay-windrowed, oats, soybeans-no\_till, soybeans-min\_till, soybeans-clean, wheat, woods, buildings-grass-trees-drives, stone-steel\_towers, separately.}
It can be noted that the developed method obtains a better performance than that by SVM. More importantly, the learned CNN by the statistical loss achieves an accuracy of 99.51\% $\pm$ 0.09\% over Pavia University which is much higher than that by general softmax loss (98.61\% $\pm$ 0.35\%). Besides, for Indian Pines, the proposed method can decrease the error rate by 47.42\% when compared with that by general softmax loss. { The statistical loss can take advantage of the statistical property of the hyperspectral image and embed the information of class distributions of the hyperspectral image in the deep learning process. Thus, the learned deep model can better represent the hyperspectral image and further provide a better classification performance.}

Furthermore, we use the McNemar's test, which is based upon the standardized normal test statistics \cite{13}, as the statistical analysis method to demonstrate whether the developed statistical loss method improve the classification performance in the statistic sense. The statistic is computed by
\begin{equation}\label{eq:35}
  F_{ij}=\frac{f_{ij}-f_{ji}}{\sqrt{f_{ij}+f_{ji}}},
\end{equation}
where $F_{ij}$ measures the pairwise statistical significance of difference between the accuracies of the $i$th and $j$th methods, and $f_{ij}$ denotes the number of samples which is classified correctly by $i$th method but wrongly by $j$th method. At the 95\% level of confidence, the difference of accuracies between different methods is statistically significant if $|F_{ij}|>1.96$.

From tables \ref{table:pavia} and \ref{table:indian}, we can find that $|F_{ij}|$ obtains 15.50 and 4.48 over Pavia University and Indian Pines, respectively, which means that the improvement of the performance by the developed statistical loss is statistically significant.

\subsection{Effects of Different Number of Training Samples}
\begin{figure}[t]
\centering
 \subfigure[]{\label{fig:number_pavia}\includegraphics[width=0.49\linewidth]{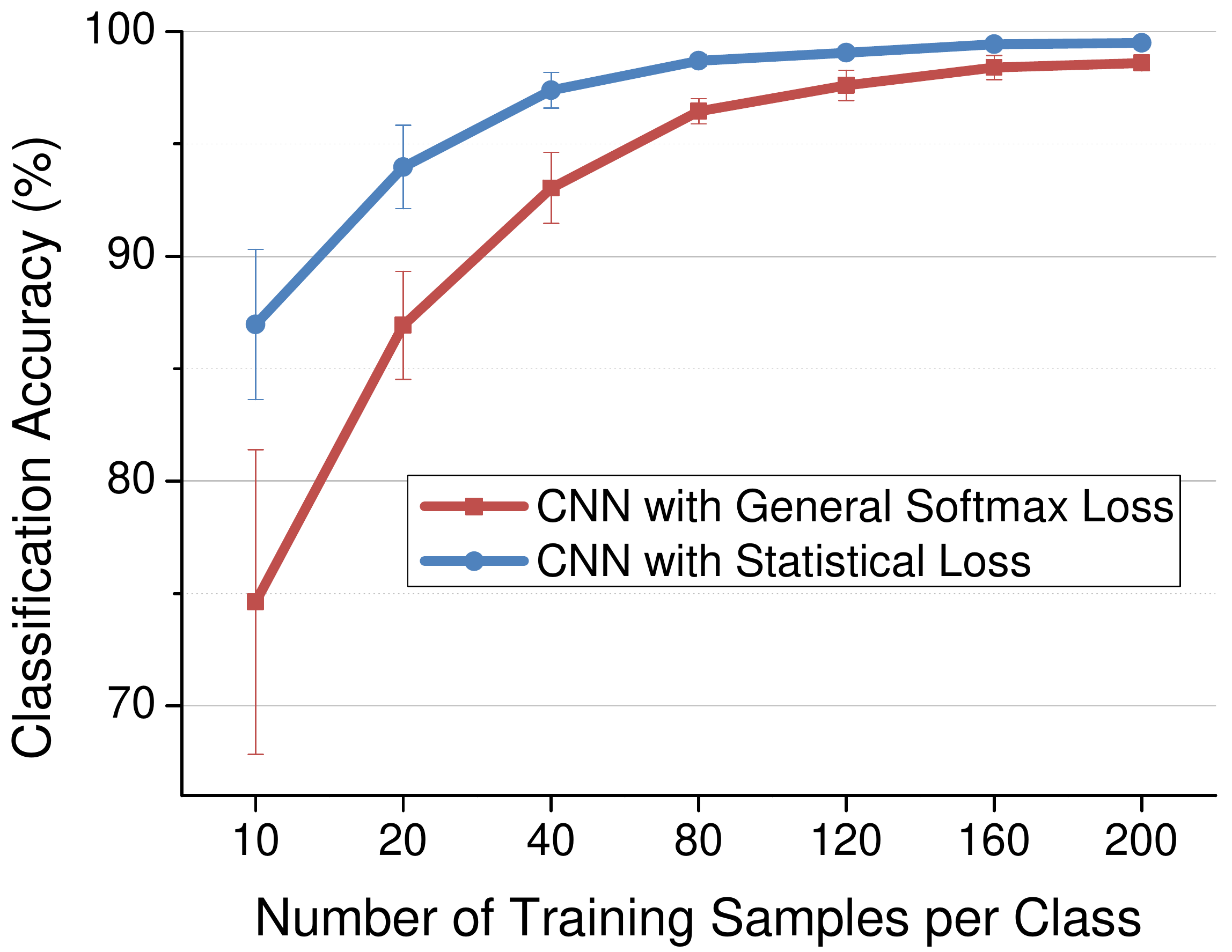}}
 \subfigure[]{\label{fig:number_indian}\includegraphics[width=0.49\linewidth]{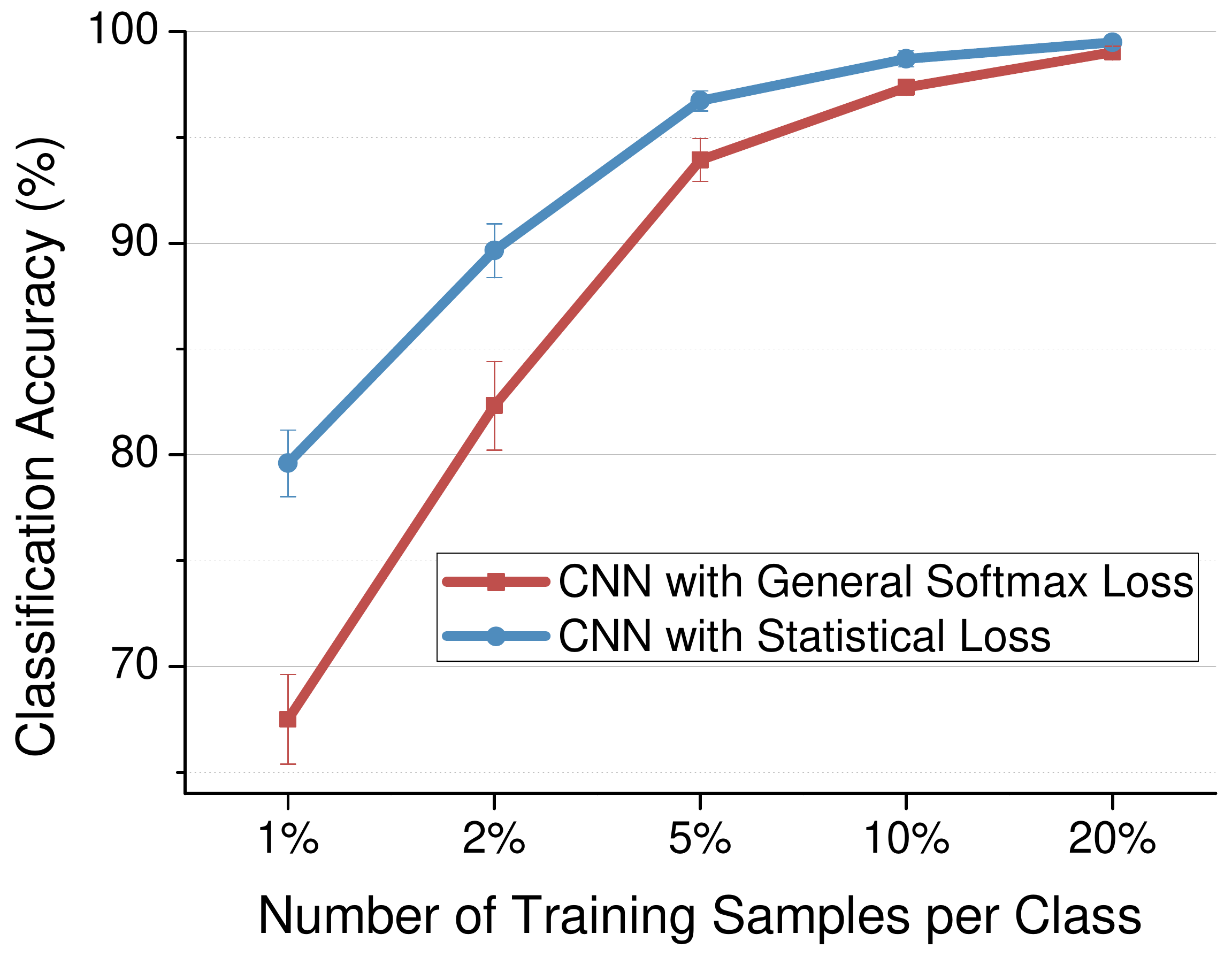}}
\vspace{-1ex}
   \caption{Classification performance with different number of training samples per class over (a) Pavia University; (b) Indian pines.}
\label{fig:performance}
\vspace{-2ex}
\end{figure}

\begin{figure}[t]
\centering
 \subfigure[]{\label{fig:mcnemar_pavia}\includegraphics[width=0.49\linewidth]{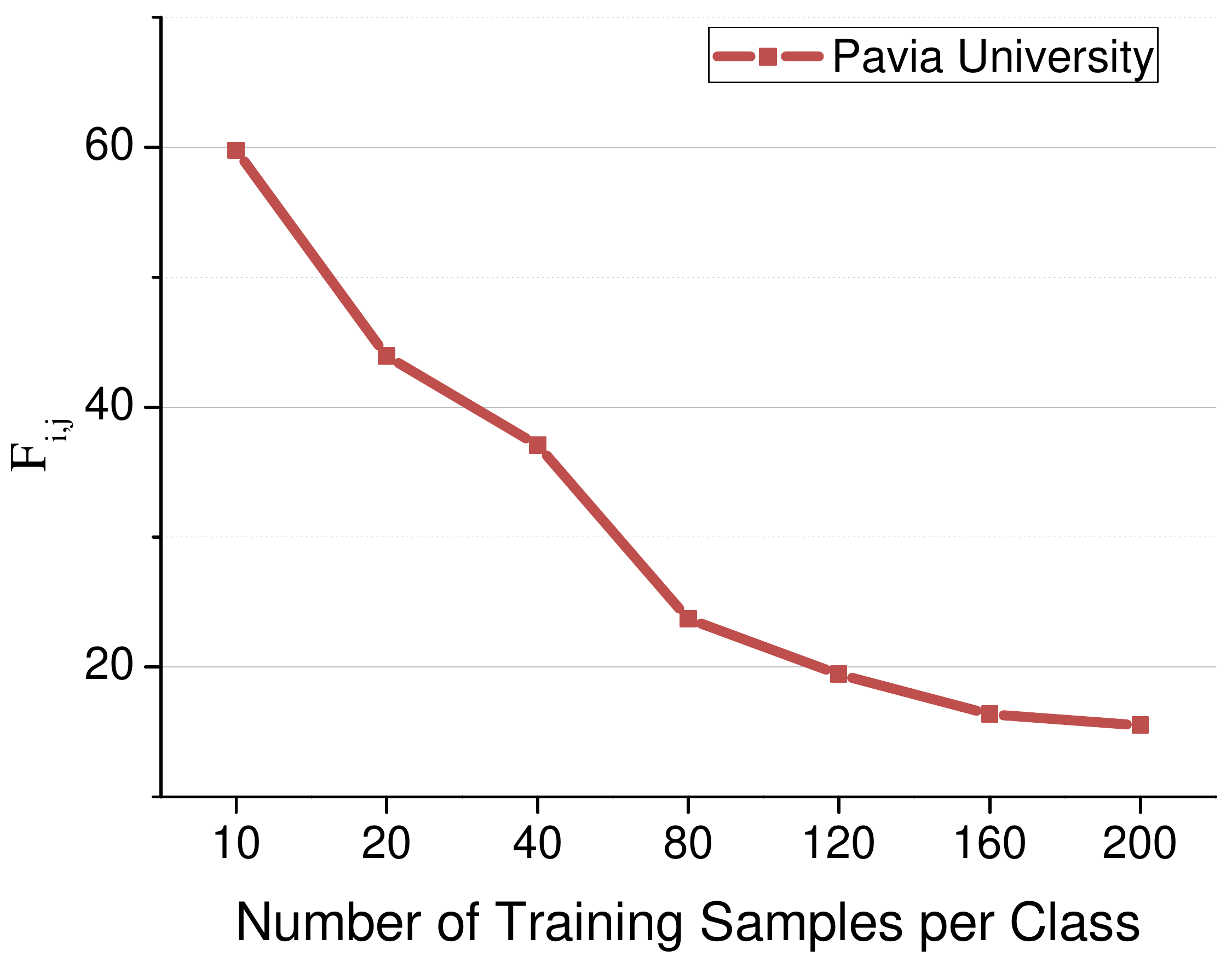}}
 \subfigure[]{\label{fig:mcnemar_indian}\includegraphics[width=0.49\linewidth]{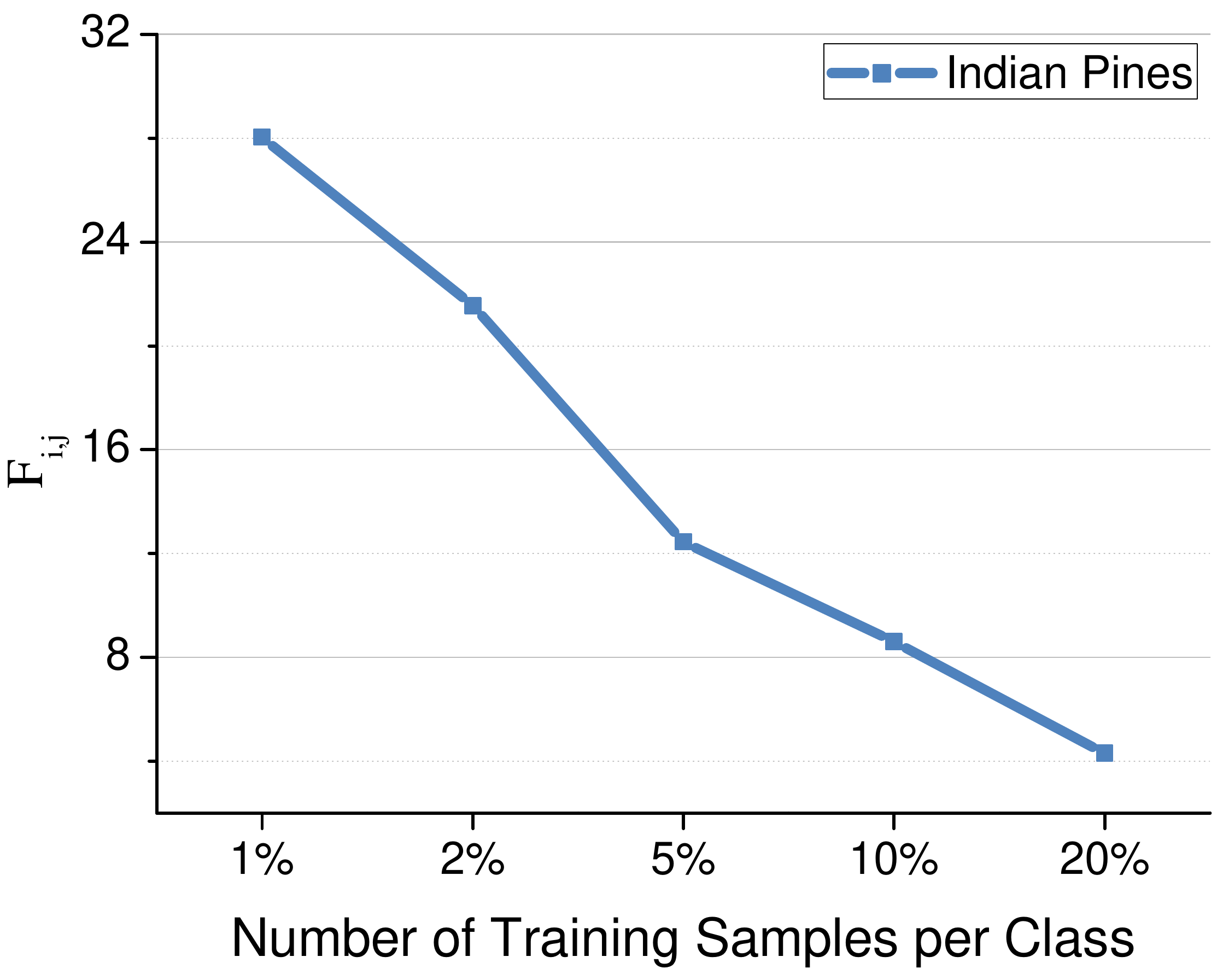}}
\vspace{-1ex}
   \caption{The Mcnemar's test between the general softmax loss and the proposed method under different number of training samples over (a) Pavia University; (b) Indian pines.}
\label{fig:mcnemar}
\vspace{-2ex}
\end{figure}

The former subsection has demonstrated the effectiveness of the developed statistical loss for hypersperctral image at the given experimental setups as section \ref{subsec:setup} shows. This subsection will further evaluate the performance of the developed method under different number of training samples. For Pavia University data, we choose the number of training samples per class from the set of $\{10,20,40,80,120,160,200\}$. While for Indian Pines data, we select 1\%, 2\%, 5\%, 10\%, and 20\% of training samples per class from the whole data. It should be noted that in these experiments, the diversity weight $\lambda$ is set to 0.01. Fig. \ref{fig:performance} presents the classification performance of the developed method with different number of training samples over the two data. Furthermore, we have presented the value of McNemar's test with different number of training samples between the CNN trained with general softmax loss and the statistical loss in Fig. \ref{fig:mcnemar}. Inspect the tendencies in Figs. \ref{fig:performance} and \ref{fig:mcnemar} and we can note that the following hold.

Firstly, the accuracies obtained by CNN with proposed method can be remarkably improved when compared with CNN by general softmax loss only. From Fig. \ref{fig:mcnemar}, we can find that all the improvement by the developed method is statistically significant when compared with general softmax loss. Particularly, the accuracy is increased from 74.62\% to 86.97\% under 10 training samples per class over Pavia University and from 67.50\% to 79.60\% under 1\% of training samples per class over Indian Pines. Secondly, the classification performance of the learned model is significantly improved with the increase of the training samples. Finally, it can be noted  that the developed statistical loss shows a definite improvement of the learned model with limited number of training samples. As showed in Fig. \ref{fig:mcnemar}, the value of MeNemar's test is significantly improved when decreasing the training samples. The $|F_{ij}|$ can even rank 59.74 under 10 training samples per class over Pavia University and 28.03 under 1\% training samples per class over Indian Pines.
{ The statistical loss is constructed with the class distributions, not directly with the samples. Therefore, even under limited training samples, the statistical loss can learn more class information with the class distributions and provide a deeply improvement of classification performance.
This indicates that the proposed method provides another way to train an effective CNN model with limited training samples.}

\begin{figure*}[t]
\centering
 \subfigure[]{\label{fig:6a}\includegraphics[width=0.14\linewidth]{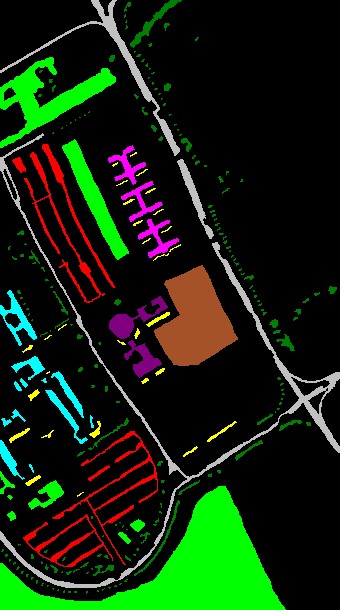}}
 \subfigure[]{\label{fig:6b}\includegraphics[width=0.14\linewidth]{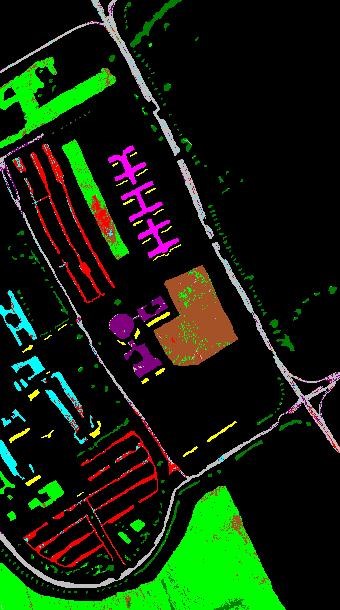}}
 \subfigure[]{\label{fig:6c}\includegraphics[width=0.14\linewidth]{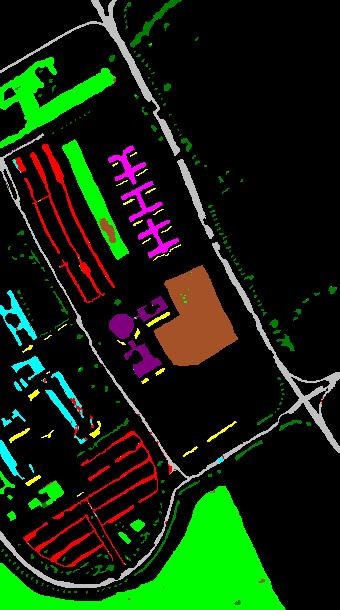}}
 \subfigure[]{\label{fig:6d}\includegraphics[width=0.14\linewidth]{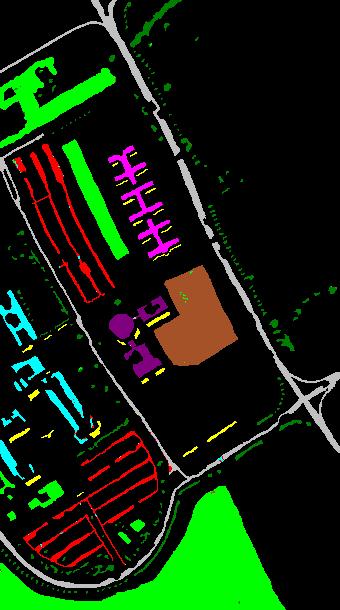}}
 \subfigure[]{\label{fig:6e}\includegraphics[width=0.14\linewidth]{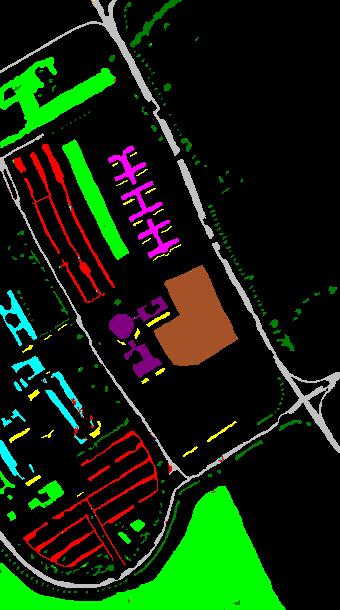}}
 \subfigure[]{\label{fig:6f}\includegraphics[width=0.14\linewidth]{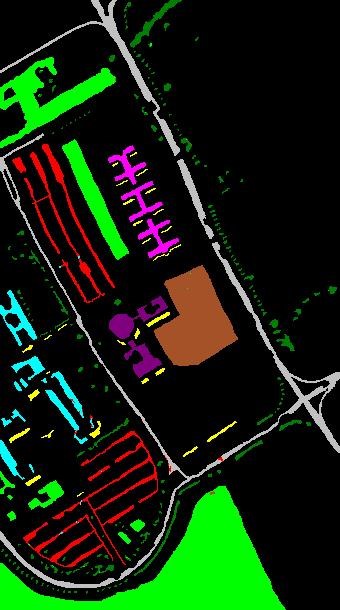}}
% \subfigure[]{\label{fig:6g}\includegraphics[width=0.107\linewidth]{6g.jpg}}
% \subfigure[]{\label{fig:6h}\includegraphics[width=0.107\linewidth]{data/6h.jpg}}
 \subfigure[]{\includegraphics[width=0.1\linewidth]{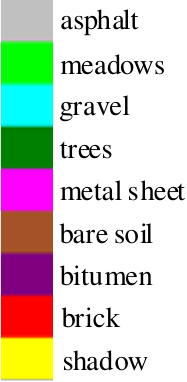}}
\vspace{-1ex}
   \caption{Pavia University classification maps by different methods with 200 samples per class for training (overall accuracies). (a) groundtruth; (b) SVM (89.2\%); (c) CNN with softmax loss (98.25\%); (d) CNN with center loss (99.44\%) ; (e) CNN with structured loss (99.25\%); (f) CNN with developed statistical loss (99.64\%); (g) map color.}
\label{fig:pavia_map}
\vspace{-2ex}
\end{figure*}

\begin{figure*}[t]
\centering
 \subfigure[]{\label{fig:7a}\includegraphics[width=0.16\linewidth]{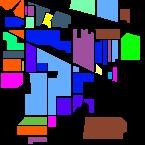}}
 \subfigure[]{\label{fig:7b}\includegraphics[width=0.16\linewidth]{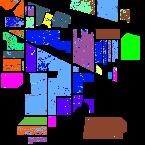}}
 \subfigure[]{\label{fig:7c}\includegraphics[width=0.16\linewidth]{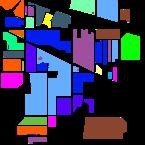}}
 \subfigure[]{\label{fig:7d}\includegraphics[width=0.16\linewidth]{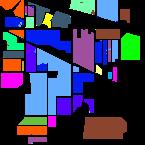}}
 \subfigure[]{\label{fig:7e}\includegraphics[width=0.16\linewidth]{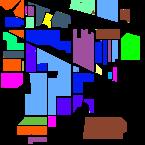}}
 \subfigure[]{\label{fig:7f}\includegraphics[width=0.16\linewidth]{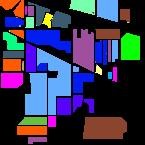}}
 \subfigure[]{\includegraphics[width=0.9\linewidth]{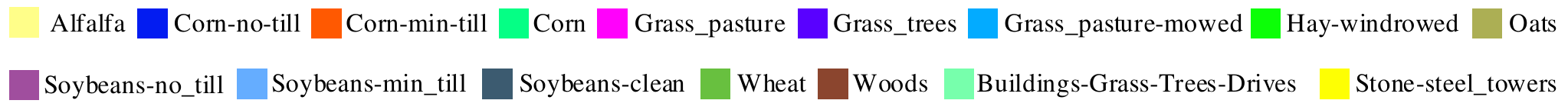}}
\vspace{-1ex}
   \caption{Indian Pines classification maps by different methods with 20\% of samples per class for training (overall accuracies). (a) groundtruth;  (b) SVM (88.15\%); (c) CNN with softmax loss (98.87\%); (d) CNN with center loss (98.91\%); (e) CNN with structured loss (99.31\%); (f) CNN with developed statistical loss (99.48\%); (g) map color.}
\label{fig:indian_map}
\vspace{-2ex}
\end{figure*}

Furthermore, we show the classification maps from different methods under 200 training samples per class over Pavia University data and 20\% of training samples per class over Indian Pines  in Figs. \ref{fig:pavia_map} and \ref{fig:indian_map}, respectively. Compare Fig. \ref{fig:6c} with \ref{fig:6f}, and \ref{fig:7c} with \ref{fig:7f}. We can find that with the statistical loss, the classification errors can be remarkably decreased over both the datasets. Besides, when compare Fig. \ref{fig:6b} with \ref{fig:6f}, and \ref{fig:7b} with \ref{fig:7f}, it can be noted that, the developed method can learn the model that is more fit for the image than general handcrafted features.

\subsection{Effects of Diversity Weight $\lambda$}
\begin{figure}[t]
\centering
 \subfigure[]{\label{fig:diversity_pavia}\includegraphics[width=0.49\linewidth]{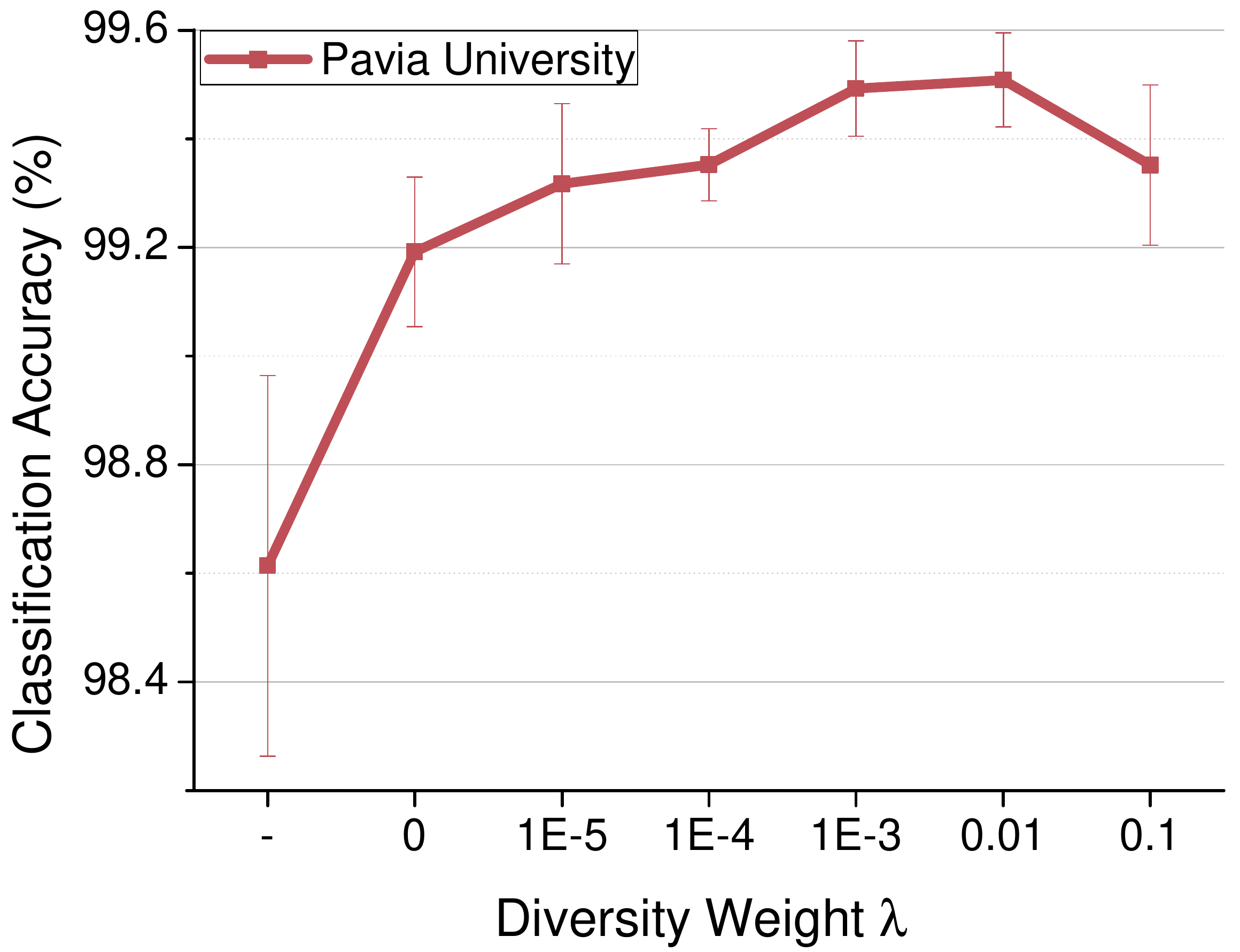}}
 \subfigure[]{\label{fig:diversity_indian}\includegraphics[width=0.49\linewidth]{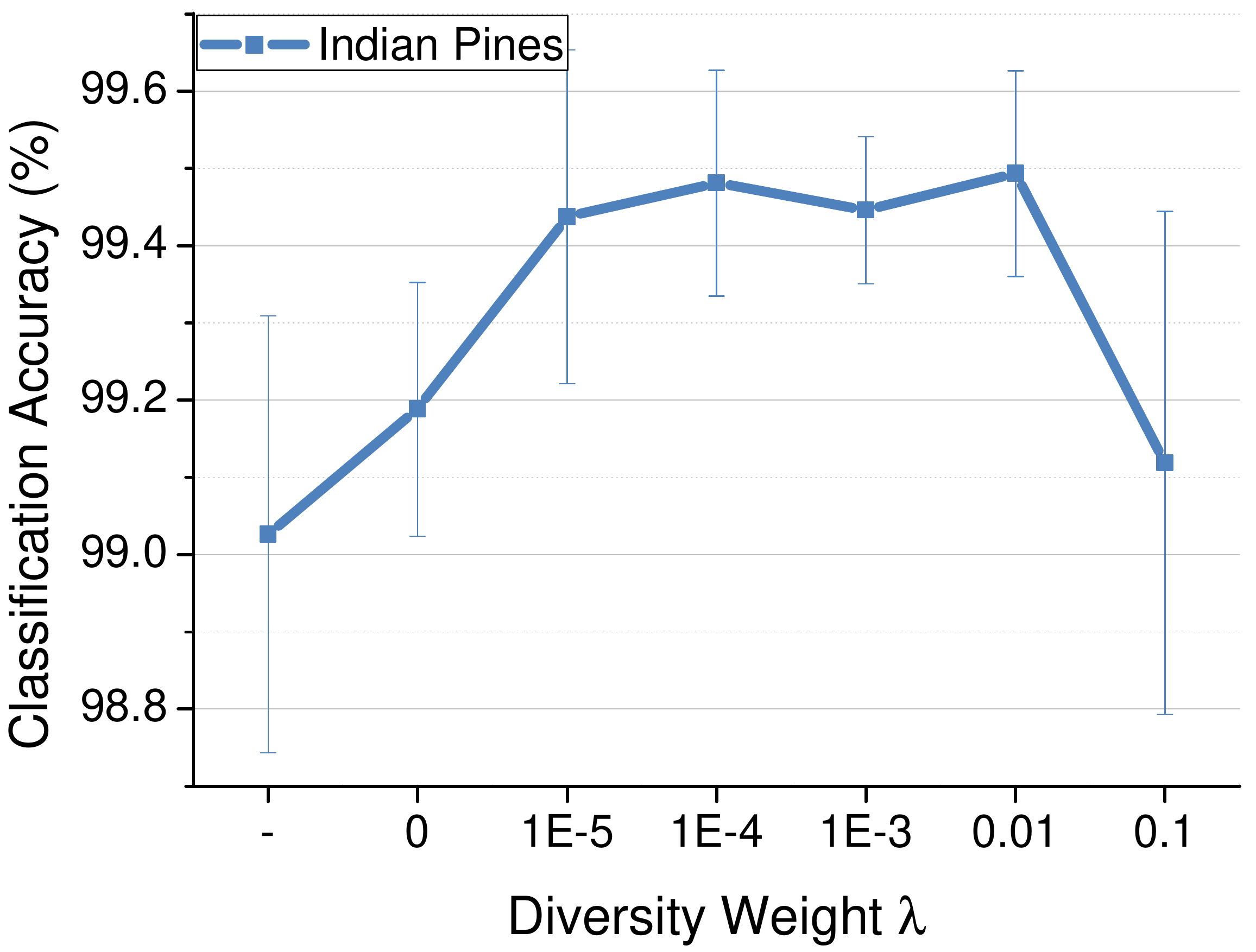}}
\vspace{-3ex}
   \caption{Classification performance of the proposed method with different diversity weight $\lambda$  over (a) Pavia University; (b) Indian pines. '$-$' represents the results obtained with general softmax loss only.}
\label{fig:diversity}
\vspace{-2ex}
\end{figure}

As mentioned in Section \ref{subsec:statistical_loss}, $\lambda$ represents the tradeoff parameter between the optimization term and the diversity term. The value of $\lambda$ can also affect the performance of the developed statistical loss. In this set of experiments, we evaluate the performance of the proposed method with different values of $\lambda$. Fig. \ref{fig:diversity} shows the classification performance with different $\lambda$ over the Pavia University and Indian Pines data, respectively.

We can find that  the statistical loss can provide a better performance with a larger $\lambda$. However, an extensively large $\lambda$ shows negatively effects on the performance of the statistical loss. Generally, increasing the $\lambda$ can encourage different class distributions to repulse from each other, and therefore, the learned features can be more discriminative to separate different objects. However, an excessively large $\lambda$ focuses too much attention on the diversity among different classes while ignores the variance of each class distribution. This could make the increase the intra-class variance of each class and show negative effects on the classification performance. More importantly, From Fig. \ref{fig:diversity}, it can be noted that the proposed method performs the best (99.51\%) when $\lambda$ is set to 0.01 over the Pavia University data. While for Indian Pines data, the accuracy ranks 99.49\% when $\lambda=0.01$ which performs the best. In practice, cross validation can be used to select a proper $\lambda$ to satisfy specific requirements of the developed statistical loss over different datasets.

\subsection{Comparisons with other Samples-based Loss}

This work also compares the developed statistical loss with other recent samples-based loss. This work selects the center loss \cite{31} and the structured loss \cite{30} as the benchmarks to characterize the pair-wise correlation between the training samples. Table \ref{table:comparison_sample_class} shows the comparison results over the Pavia University and the Indian Pines data, respectively.

From the table, we can find that the developed statistical loss which  formulates the penalization with the class distributions can be more fit for the classification task than the center loss and the structured loss. { Using 200 samples per class for training}, for Pavia Unviersity data, the statistical loss achieves $99.51\%\pm 0.09\%$ OA outperforms that by the center loss ($99.28\%\pm 0.11\%$ OA) and the structured loss ($99.27\% \pm 0.12\%$ OA). While for the Indian Pines data, it can obtain $99.49\% \pm 0.13\%$ OA { with 20\% training samples} which is higher than $99.23\% \pm 0.21\%$ OA by the center loss and $99.23\% \pm 0.25\%$ OA by the structured loss. Moreover, it can also be noted that the $|F_{ij}|$ also achieves 5.52 and 5.68 when compared with the center loss and structured loss over the Pavia University. Besides, the $|F_{ij}|$ also obtains 2.64 and 3.56 over the Indian Pines. This means that by Mcnemar's test, the developed statistical loss is statistically significant than other samples-based loss.

{ Besides, compare the statistical loss with these samples-based losses under limited number of training samples and we can also find that the deep model can obtain a significant improvement with the developed method. The reason is that the statistical loss is constructed with the class distributions and can use more class information in the training process while the samples-based losses are constructed directly with the training samples.
In conclusion, the developed statistical loss which is formulated with the class distributions can achieve superior performance when compared with other samples-based loss in the literature of hyperspectral image classification.}

The classification maps from CNN model learned with the center loss, the structured loss over the two datasets are shown in Figs. \ref{fig:pavia_map} and \ref{fig:indian_map}, separately. Compare Fig. \ref{fig:6d} with Fig. \ref{fig:6f} and Fig. \ref{fig:7d} with Fig. \ref{fig:7f}, and { it can be easily found that the CNN model with the statistical loss can better model the hyperspectral image than that with the center loss.} Besides, compare Fig. \ref{fig:6e} with Fig. \ref{fig:6f} and Fig. \ref{fig:7e} with Fig. \ref{fig:7f} and we can obtain that the statistical loss can significantly decrease the classification errors than that by the structured loss.

\begin{table*}[t]
\begin{center}
\caption{Comparisons with other sample-wise loss. This work selects the generally used softmax loss. Furthermore, this work compares the developed statistical loss with other most recent sample-based loss, namely the center loss \cite{31} and the structured loss \cite{30}. PU represents the Pavia University and IP represents the Indian Pines data.}
\label{table:comparison_sample_class}
\vspace{-2ex}
\begin{tabular}{| c | c | c | c | c | c | c |}
%\toprule[1pt]
\hline
{\bf Data}    &   {\bf Training set} &   {\bf Methods}     &  {\bf OA(\%)} &  {\bf AA(\%)} &  {\bf KAPPA(\%)} & $F_{ij}$ \\
\hline\hline
\multirow{12}{*}{\textbf{PU}}& \multirow{4}{*}{\textbf{10 per class}}& {\bf Softmax Loss}    &  $74.62 \pm 6.79 $ &  ${81.81 \pm 3.38 }$  &  ${67.98 \pm 7.90 }$  & 59.74  \\
& & {\bf Center Loss}    &  $ 83.18 \pm 3.26$ &  $ 88.36 \pm 2.28 $  &  $ 79.34 \pm 4.72 $  & 27.17 \\
& & {\bf Structured Loss} &  $ 81.03 \pm 5.05$ &  $87.03 \pm 2.31 $  &  $75.99 \pm 5.80 $  & 37.84 \\
& & {\bf Statistical Loss } &  $\mathbf{ 86.97 \pm 3.34 }$ &  $\mathbf{ 91.25 \pm 0.77 }$  &  $\mathbf{ 83.29 \pm 3.98 }$  & $-$ \\
\cline{2-7}
&\multirow{4}{*}{\textbf{20 per class}} & {\bf Softmax Loss}    &  $ 86.93 \pm 2.41 $ &  $ 89.82 \pm 1.29 $  &  $ 83.11 \pm 2.84 $  & 43.92  \\
& & {\bf Center Loss}    &  $ 92.32 \pm 2.85 $ &  $ 94.08 \pm 0.67 $  &  $ 89.99 \pm 3.55 $  & 21.81  \\
& & {\bf Structured Loss} &  $89.80 \pm 3.39 $ &  $ 93.20 \pm 0.88 $  &  $ 86.84 \pm 4.18 $  & 31.28 \\
& & {\bf Statistical Loss } &  $\mathbf{ 93.98 \pm 1.86 }$ &  $\mathbf{95.38 \pm 0.73 }$  &  $\mathbf{92.12 \pm 2.37 }$  & $-$ \\
\cline{2-7}
& \multirow{4}{*}{\textbf{200 per class}}& {\bf Softmax Loss}    &  $98.61 \pm 0.35$ &  ${98.56 \pm 0.36}$  &  ${98.14 \pm 0.47}$  & 15.50\\
& & {\bf Center Loss}    &  $99.28 \pm 0.11$ &  $99.13 \pm 0.17$  &  $99.03 \pm 0.14$  & 5.52 \\
& & {\bf Structured Loss} &  $99.27 \pm 0.12$ &  $99.12 \pm 0.22$  &  $99.02 \pm 0.17$  & 5.68 \\
& & {\bf Statistical Loss } &  $\mathbf{99.51 \pm 0.09}$ &  $\mathbf{99.37 \pm 0.13}$  &  $\mathbf{99.34 \pm 0.12}$  & $-$ \\
 \hline\hline
\multirow{12}{*}{\textbf{IP}} & \multirow{4}{*}{\textbf{1\%}} &{\bf Softmax Loss}    &  $ 67.50 \pm 2.12 $ &  $ 57.36 \pm 2.45 $  &  $ 62.63 \pm 2.44 $  & 28.03 \\
& & {\bf Center Loss}    &  $ 73.25 \pm 2.84 $ &  ${64.20 \pm 4.69 }$  &  ${ 69.37 \pm 3.26 }$  & 16.97 \\
& &{\bf Structured Loss}    &  $ 71.44 \pm 1.93 $ &  ${ 61.81 \pm 3.43 }$  &  ${ 67.31 \pm 2.27 }$  & 20.93  \\
& &  {\bf Statistical Loss} &  $\mathbf{ 79.60 \pm 1.57 }$ &  $\mathbf{ 68.95 \pm 3.25 }$  &  $\mathbf{ 76.61 \pm 1.88 }$  & $-$ \\
\cline{2-7}
& \multirow{4}{*}{\textbf{2\%}} &{\bf Softmax Loss}    &  $ 82.32 \pm 2.10 $ &  $ 72.83 \pm 4.65 $  &  $ 79.77 \pm 2.41 $  & 21.54  \\
& & {\bf Center Loss}    &  $ 85.83 \pm 1.49 $ &  ${ 77.18 \pm 2.68 }$  &  ${ 83.82 \pm 1.71 }$  & 12.98  \\
& &{\bf Structured Loss}    &  $ 84.63\pm 1.66 $ &  ${ 76.35 \pm 2.46 }$  &  ${ 82.46 \pm 1.88 }$  & 16.27   \\
& &  {\bf Statistical Loss} &  $\mathbf{ 89.65 \pm 1.28 }$ &  $\mathbf{ 79.08 \pm 3.42 }$  &  $\mathbf{88.16 \pm 1.48 }$  & $-$ \\
\cline{2-7}
& \multirow{4}{*}{\textbf{20\%}} &{\bf Softmax Loss}    &  $99.03 \pm 0.28$ &  $98.79 \pm 0.45$  &  $98.89 \pm 0.32$  & 4.48 \\
& & {\bf Center Loss}    &  $99.23 \pm 0.21$ &  $\mathbf{98.97 \pm 0.42}$  &  ${99.12 \pm 0.23}$  & 2.64 \\
& &{\bf Structured Loss}    &  $99.13 \pm 0.25$ &  ${98.75 \pm 0.35}$  &  ${99.01 \pm 0.29}$  &  3.56 \\
& &  {\bf Statistical Loss} &  $\mathbf{99.49 \pm 0.13}$ &  ${98.40 \pm 0.93}$  &  $\mathbf{99.42 \pm 0.15}$  & $-$ \\
\hline

%\bottomrule[1pt]
\end{tabular}
\end{center}
\vspace{-2ex}
\end{table*}

\subsection{Comparisons with the Most Recent Methods}

To further validate the effectiveness of the developed statistical loss for hyperspectral image classification, we compare the developed statistical loss with the state-of-the-art methods. Tables III and IV show the comparisons over the two datasets, respectively. The experimental results in each table are with the same experimental setups and we use the results from the literature where the method is first developed directly.

\begin{table*}[t]
\begin{center}
\caption{Classification performance of different methods over Pavia Unviersity data in the most recent literature (200 training samples per class for training). }
\label{table:comparison_pavia}
\vspace{-1ex}
\begin{tabular}{| c | c | c | c |}
%\toprule[1pt]
\hline
{\bf Methods}     &  {\bf OA(\%)} &  {\bf AA(\%)} &  {\bf KAPPA(\%)} \\
\hline\hline

{\bf SVM-POLY} &  $88.07\pm 0.82$  &  $91.53\pm 0.26$ &  ${84.35\pm 1.01}$\\
{\bf D-DBN-PF \cite{17}}    &  $93.11\pm 0.06$ &  ${93.92\pm 0.07}$  &  ${90.82\pm 0.10}$\\
{\bf CNN-PPF \cite{15}}    &  $96.48$ &  $-$  &  $-$\\
 {\bf Contextual DCNN \cite{04}} &  $97.31\pm 0.26$ &  $-$  &  $-$\\
{\bf SSN \cite{14}}    &  $99.36\pm 0.11$ &  $-$  &  $-$\\
  {\bf ML-based Spec-Spat \cite{19}}    &  $99.34$ &  ${99.40}$  &  ${99.11}$\\
{\bf DPP-DML-MS-CNN \cite{01}}    &  $99.46\pm 0.03$ &  ${99.39\pm 0.05}$  &  ${99.27\pm 0.04}$\\
  {\bf Proposed Method} &  $\mathbf{99.51\pm 0.09}$ &  $\mathbf{99.37\pm 0.13}$  &  $\mathbf{99.34\pm 0.12}$\\
\hline

%\bottomrule[1pt]
\end{tabular}
\end{center}
\vspace{-2ex}
\end{table*}

\begin{table*}[t]
\begin{center}
\caption{Classification performance of different methods over Indian Pines data in the most recent literature. In the table, the results of \cite{46} come from the literature \cite{45}. The percent in the brackets demonstrates the training samples per class.}
\label{table:comparison_indian}
\vspace{-1ex}
\begin{tabular}{| c | c | c | c |}
%\toprule[1pt]
\hline
{\bf Methods}     &  {\bf OA(\%)} &  {\bf AA(\%)} &  {\bf KAPPA(\%)} \\
\hline\hline

{\bf R-ELM \cite{09}} &  $97.62$  &  $97.26$ &  ${97.29}$\\
{\bf DEFN \cite{45}}    &  $98.52\pm 0.23$ &  ${98.32\pm 0.26}$  &  ${97.69\pm 0.74}$\\
{\bf DRN \cite{46}}    &  $98.36\pm 0.42$ &  $98.13\pm 0.48$  &  $97.62\pm 0.79$\\
 {\bf MCMs+2DCNN \cite{47}} &  $98.61\pm 0.30$ &  $96.94\pm 0.78$  &  $98.42\pm 0.34$\\
 {\bf Proposed Method (10\%) } &  $\mathbf{98.72\pm 0.40}$ &  $\mathbf{95.17\pm 2.01}$  &  $\mathbf{98.54\pm 0.43}$\\
 \hline\hline
{\bf SVM-POLY}    &  $88.20\pm 0.51$ &  $85.05\pm 1.26$  &  $86.49\pm 0.58$\\
  {\bf SSRN \cite{20}}    &  $99.19\pm 0.26$ &  ${98.93\pm 0.59}$  &  ${99.07\pm 0.30}$\\
{\bf MCMs+2DCNN \cite{47}}    &  $99.07\pm 0.25$ &  ${99.04\pm 0.27}$  &  ${98.94\pm 0.29}$\\
  {\bf Proposed Method (20\%)} &  $\mathbf{99.49\pm 0.13}$ &  $\mathbf{98.40\pm 0.93}$  &  $\mathbf{99.42\pm 0.15}$\\
\hline

%\bottomrule[1pt]
\end{tabular}
\end{center}
\vspace{-2ex}
\end{table*}

From table \ref{table:comparison_pavia}, we can obtain that the proposed method which can obtain 99.51\% $\pm$ 0.09\% OA outperforms the D-DBN-PF (93.11\% $\pm$ 0.06\% OA) \cite{17}, CNN-PPF (96.48\%) \cite{15}, Contextual DCNN (97.31\% $\pm$ 0.26\% OA) \cite{04}, SSN (99.36\% $\pm$ 0.11\% OA) \cite{14}, ML-based Spec-Spat (99.34\% OA) \cite{19}, and DPP-DML-MS-CNN (99.46\% $\pm$ 0.03\% OA) \cite{01} over Pavia University. As listed in table \ref{table:comparison_indian}, for Indian Pines data, when using 10\% of samples per class for training, the proposed method which can obtain 98.72\% $\pm$ 0.40\% OA outperforms the R-ELM (97.62\% OA) \cite{09}, DEFN (98.52\% $\pm$ 0.23\% OA) \cite{45}, DRN (98.36\% $\pm$ 0.42\% OA) \cite{46}, and MCMs+2DCNN (98.61\% $\pm$ 0.30\%) \cite{47}. Moreover, when using 20\% of samples per class for training, the accuracy can achieve 99.49\% $\pm$ 0.13\% OA by the proposed method which is higher than 99.19\% $\pm$ 0.26\% OA by SSRN \cite{20} and 99.07\% $\pm$ 0.25\% OA by MCMs+2DCNN \cite{47} over Indian Pines data. Overall, the proposed method which takes advantage of the statistical properties of the hyperspectral image and formulates the penalization with the class distributions can obtain a
comparable or even better performance when compared with other state-of-the-art methods over the hyperspectral image classification.

%Besides, we also compare the results of the developed method with the representative methods which focus on the classification task with limited training samples. Here, we mainly choose \cite{48} and \cite{49} as baselines. The proposed method in \cite{48} can achieve 94.67\% $\pm$ 0.25\% OA with 5\% of training samples per class over Indian Pines while the proposed method in this work can obtain 96.73\% $\pm$ 0.48\% OA with the same experimental setup. Besides, the proposed method can obtain 86.97\% $\pm$ 3.34\% OA with 10 samples per class over Pavia University which is better than that by NFE with 15 samples per class (81.99\% $\pm$ 1.42\% OA). From the comparisons with these methods, it can be noted that the developed method can be effective for tasks with limited training samples and our proposed method can provide another choices for these deep learning tasks with only limited training samples.

\section{Conclusions}
In this work, based on the statistical properties of the hyperspectral image and multi-variant statistical analysis, we develop a novel statistical loss for  hyperspectral image classification. First, we characterize each class from the image as a specific probabilistic model. Then, according to the Fisher discrimination criterion, we develop the statistical loss with distributions for deep learning. The experimental results have shown the effectiveness of the proposed method when compared with other most recent samples-based loss as well as the state-of-the-art methods in hyperspectral image classification.
%Furthermore, the results  validated that the proposed method can be effective in the deep learning tasks with limited training samples.

In future works, it would be interesting to investigate the performance of the developed statistical loss on other CNN model. Besides, investigating the effects of the proposed statistical loss on the applications of other computer vision tasks is an important future topic.

\appendix[Supplemental Materials]

\section{Overview}

% If you have more than one objective, uncomment the below:
%\begin{description}
%\item[First Objective] \hfill \\
%Objective 1 text
%\item[Second Objective] \hfill \\
%Objective 2 text
%\end{description}
This part includes supplementary material to ``Statistical Loss and Analysis for Deep Learning in Hyperspectral Image Classification''. Included are detailed versions of the algorithms and other more results.

%----------------------------------------------------------------------------------------
%	SECTION 2
%----------------------------------------------------------------------------------------
\section{Computation of Gradients}

\subsection{Some Formulations about the Gradients of the Matrix}

{\bf Gradients of the product of Matrix.} The gradients of the product of vectors satisfy the chain rule. Denote ${\bf a},{\bf b},{\bf x}$ as $p\times 1$ vectors, then
\begin{equation}\label{eq:f02}
  \frac{\partial {\bf a}^T{\bf b}}{\partial {\bf x}}=\frac{\partial {\bf b}^T}{\partial {\bf x}}{\bf a}+\frac{\partial {\bf a}^T}{\partial {\bf x}}{\bf b}.
\end{equation}
Denote $A, B$ as matrices and $y$ as a scalar, then
\begin{equation}\label{eq:f003}
  \frac{\partial AB}{\partial y}=\frac{\partial A}{\partial y}B+A\frac{\partial B}{\partial y}.
\end{equation}

{\bf Gradients of the inverse of a certain matrix.} Denote $A$ as a certain matrix and $y$ as a scalar, then
\begin{equation}\label{eq:f01}
  \frac{\partial A^{-1}}{\partial y}=-A^{-1}\frac{\partial A}{\partial y}A^{-1}
\end{equation}
\begin{proof}
\begin{equation*}
  AA^{-1}=I_0,
\end{equation*}
where $I_0$ denotes the identity matrix. Then
\begin{equation*}
  \frac{\partial AA^{-1}}{\partial y}=0,
\end{equation*}
Based on the chain rule,
\begin{equation*}
  \frac{\partial A}{\partial y}A^{-1}+A\frac{\partial A^{-1}}{\partial y}=0
\end{equation*}
Therefore,
\begin{equation*}
  \frac{\partial A^{-1}}{\partial y}=-A^{-1}\frac{\partial A}{\partial y}A^{-1}
\end{equation*}
\end{proof}

\subsection{Gradients of the Statistical Loss}
%For conciseness, only the computations of the gradient of the developed statistical loss will be presented.
The final objective functions of the proposed statistical loss for hyperspectral image classification can be finally written as follows:
\begin{equation}\label{eq:f03}
  L=L_0+\lambda L_{div}
\end{equation}
where
\begin{align}
\label{eq:04}  L_0=\frac{1}{\Lambda}\sum_{k=1}^{\Lambda}(\frac{1}{n_k-1}\sum_{j=1}^{n_k}({\bf z}_j-\overline{C}_k)^T({\bf z}_j-\overline{C}_k)) \\
\label{eq:05}  L_{div}=\sum_{k\neq t}^{\Lambda}(\Delta-\frac{n_k+n_t-2}{\frac{1}{n_k}+\frac{1}{n_t}}\Gamma_{k,t}^T(S_k+S_t)^{-1}\Gamma_{k,t})
\end{align}

The computations of the gradients of the terms are provided below.

{\bf Computing} ${\displaystyle{\frac{\partial L_0}{\partial {\bf z}_i}}}$. Based on Eq. \ref{eq:f02}, the gradients can be easily computed by
\begin{equation}\label{eq:f06}
  \frac{\partial L_0}{\partial {\bf z}_i}=\frac{2}{\Lambda}\sum_{k=1}^{\Lambda}\frac{1}{n_k}I({\bf z}_i\in C_k^B)({\bf z}_i-\overline{C}_k)
\end{equation}
{ where $I({\text{condition}})$ denotes the indicative function}. $I(\cdot)=1$ if the condition is true and $I(\cdot)=0$ if not.

{ {\bf Computing} $\displaystyle{ \frac{\partial L_{div}}{\partial {\bf z}_i}}$}. The partial of $L_{div}$ w.r.t. ${\bf z}_i$ can be calculated by
\begin{equation}\label{eq:f07}
  \frac{\partial L_{div}}{\partial {\bf z}_i}=-\sum_{k\neq t}^{\Lambda}\frac{n_k+n_t-2}{\frac{1}{n_k}+\frac{1}{n_t}}\frac{\partial \Gamma_{k,t}^T(S_k+S_t)^{-1}\Gamma_{k,t}}{\partial {\bf z}_i}
\end{equation}
Therefore, the main problem is to calculate the { derivation} of $\displaystyle{\frac{\partial \Gamma_{k,t}^T(S_k+S_t)^{-1}\Gamma_{k,t}}{\partial {\bf z}_i}}$. Based on Eq. \ref{eq:f02}, it can be reformulated as
\begin{equation}\label{eq:f08}
\begin{aligned}
  \frac{\partial \Gamma_{k,t}^T(S_k+S_t)^{-1}\Gamma_{k,t}}{\partial {\bf z}_i}=&\frac{\partial \Gamma_{k,t}^T(S_k+S_t)^{-1}}{\partial {\bf z}_i}\Gamma_{k,t}  \\
  &+\frac{\partial \Gamma_{k,t}^T}{\partial {\bf z}_i}(S_k+S_t)^{-1}\Gamma_{k,t}
  \end{aligned}
\end{equation}
By the definition of the { derivation} of a vector w.r.t. a certain vector, it can easily obtain that
\begin{equation}\label{eq:f09}
  \frac{\partial \Gamma_{k,t}^T}{\partial {\bf z}_i}=\frac{\partial (\overline{C}_k-\overline{C}_t)^T}{\partial {\bf z}_i}=\frac{1}{n_k}I({\bf z}_i\in C_k)I_0
\end{equation}
where $I_0$ represents the identity matrix.

Furthermore, define ${\bf z}_i=[z_{i1},z_{i2},\cdots,z_{ip}]^T$ where $p$ is the dimension of the features ${\bf z}_i$, and then { depending on Eq. \ref{eq:f003}}, $\displaystyle{\frac{\partial \Gamma_{k,t}^T(S_k+S_t)^{-1}}{\partial {\bf z}_i}}$ can be formulated as
\begin{equation}\label{eq:f10}
\begin{aligned}
&\frac{\partial \Gamma_{k,t}^T(S_k+S_t)^{-1}}{\partial {\bf z}_i}=
  \left[
  \begin{aligned}
  \frac{\partial \Gamma_{k,t}^T(S_k+S_t)^{-1}}{\partial { z}_{i1}} \\
  \frac{\partial \Gamma_{k,t}^T(S_k+S_t)^{-1}}{\partial { z}_{i2}} \\
  \vdots\ \ \ \ \ \ \ \ \ \ \\
  \frac{\partial \Gamma_{k,t}^T(S_k+S_t)^{-1}}{\partial { z}_{ip}}
  \end{aligned}
  \right]  \\
  &=
  \left[
  \begin{aligned}
  \frac{\partial \Gamma_{k,t}^T}{\partial { z}_{i1}}(S_k+S_t)^{-1}+\Gamma_{k,t}^T\frac{\partial (S_k+S_t)^{-1}}{\partial { z}_{i1}} \\
  \frac{\partial \Gamma_{k,t}^T}{\partial { z}_{i2}}(S_k+S_t)^{-1}+\Gamma_{k,t}^T\frac{\partial (S_k+S_t)^{-1}}{\partial { z}_{i2}} \\
  \vdots\ \ \ \ \ \ \ \ \ \ \ \ \ \ \ \ \ \ \ \ \ \ \ \ \\
  \frac{\partial \Gamma_{k,t}^T}{\partial { z}_{ip}}(S_k+S_t)^{-1}+\Gamma_{k,t}^T\frac{\partial (S_k+S_t)^{-1}}{\partial { z}_{ip}}
  \end{aligned}
  \right]    \\
  &=
  \left[
  \begin{aligned}
  \frac{\partial \Gamma_{k,t}^T}{\partial { z}_{i1}}(S_k+S_t)^{-1} \\
  \frac{\partial \Gamma_{k,t}^T}{\partial { z}_{i2}}(S_k+S_t)^{-1} \\
  \vdots\ \ \ \ \ \ \ \ \ \ \\
  \frac{\partial \Gamma_{k,t}^T}{\partial { z}_{ip}}(S_k+S_t)^{-1}
  \end{aligned}
  \right] +
  \left[
  \begin{aligned}
  \Gamma_{k,t}^T\frac{\partial (S_k+S_t)^{-1}}{\partial { z}_{i1}} \\
  \Gamma_{k,t}^T\frac{\partial (S_k+S_t)^{-1}}{\partial { z}_{i2}} \\
  \vdots\ \ \ \ \ \ \ \ \ \ \\
  \Gamma_{k,t}^T\frac{\partial (S_k+S_t)^{-1}}{\partial { z}_{ip}}
  \end{aligned}
  \right]
\end{aligned}
\end{equation}

{ Based on Eq. \ref{eq:f01}, $\displaystyle{\frac{\partial \Gamma_{k,t}^T(S_k+S_t)^{-1}}{\partial {\bf z}_i}}$ can be reformulated as}
\begin{equation}\label{eq:f11}
\begin{aligned}
  \frac{\partial \Gamma_{k,t}^T(S_k+S_t)^{-1}}{\partial {\bf z}_i}=&\frac{\partial \Gamma_{k,t}^T}{\partial {\bf z}_i}(S_k+S_t)^{-1} \\
  &+\left[
  \begin{aligned}
  \Gamma_{k,t}^T\frac{\partial (S_k+S_t)^{-1}}{\partial { z}_{i1}} \\
  \Gamma_{k,t}^T\frac{\partial (S_k+S_t)^{-1}}{\partial { z}_{i2}} \\
  \vdots\ \ \ \ \ \ \ \ \ \ \\
  \Gamma_{k,t}^T\frac{\partial (S_k+S_t)^{-1}}{\partial { z}_{ip}}
  \end{aligned}
  \right]   \\
=&\frac{\partial \Gamma_{k,t}^T}{\partial {\bf z}_i}(S_k+S_t)^{-1} \\
  &-
  \left[
  \begin{aligned}
  \Gamma_{k,t}^T(S_k+S_t)^{-1}\frac{\partial (S_k+S_t)}{\partial { z}_{i1}} \\
  \Gamma_{k,t}^T(S_k+S_t)^{-1}\frac{\partial (S_k+S_t)}{\partial { z}_{i2}} \\
  \vdots\ \ \ \ \ \ \ \ \ \ \\
  \Gamma_{k,t}^T(S_k+S_t)^{-1}\frac{\partial (S_k+S_t)}{\partial { z}_{ip}}
  \end{aligned}
  \right](S_k+S_t)^{-1}
  \end{aligned}
\end{equation}
Besides, $S_k=\sum_{j=1}^{n_k}({\bf z}_j-\overline{C}_k)({\bf z}_j-\overline{C}_k)^T$. Therefore,
\begin{equation}\label{eq:f12}
\begin{aligned}
  &\frac{\partial \Gamma_{k,t}^T(S_k+S_t)^{-1}}{\partial {\bf z}_i}=\frac{\partial \Gamma_{k,t}^T}{\partial {\bf z}_i}(S_k+S_t)^{-1} \\
  &-I({\bf z}_i\in C_k^B)
  \left[
  \begin{aligned}
  \Gamma_{k,t}^T(S_k+S_t)^{-1}\frac{\partial S_k}{\partial { z}_{i1}} \\
  \Gamma_{k,t}^T(S_k+S_t)^{-1}\frac{\partial S_k}{\partial { z}_{i2}} \\
  \vdots\ \ \ \ \ \ \ \ \ \ \\
  \Gamma_{k,t}^T(S_k+S_t)^{-1}\frac{\partial S_k}{\partial { z}_{ip}}
  \end{aligned}
  \right](S_k+S_t)^{-1}
  \end{aligned}
\end{equation}
Then,
\begin{equation}\label{eq:f13}
\begin{aligned}
  &\frac{\partial \Gamma_{k,t}^T(S_k+S_t)^{-1}}{\partial {\bf z}_i}=\frac{\partial \Gamma_{k,t}^T}{\partial {\bf z}_i}(S_k+S_t)^{-1}\\
  &-\frac{n_k-1}{n_k}I({\bf z}_i\in C_k^B)[(\overline{C}_k-\overline{C}_t)^T(S_k+S_t)^{-1}{\bf z}_i](S_k+S_t)^{-1} \\
  &-\frac{n_k-1}{n_k}I({\bf z}_i\in C_k^B)(S_k+S_t)^{-1}(\overline{C}_k-\overline{C}_t){\bf z}_i^T(S_k+S_t)^{-1}
\end{aligned}
\end{equation}

{ According to Eqs. \ref{eq:f08}, \ref{eq:f09}, and \ref{eq:f13}}, we can obtain the gradients of $\displaystyle{\frac{\partial L_{div}}{\partial {\bf z}_i}}$ as
\begin{equation}\label{eq:f14}
\begin{aligned}
  &\frac{\partial L_{div}}{\partial {\bf z}_i}=\frac{n_k+n_t-2}{\frac{1}{n_k}+\frac{1}{n_t}}[-\frac{2}{n_k}I({\bf z}_i\in C_k^B)(S_k+S_t)^{-1}(\overline{C}_k-\overline{C}_t) \\
  &+\frac{n_k-1}{n_k}I({\bf z}_i\in C_k^B)[(\overline{C}_k-\overline(C)_t)^T(S_k+S_t)^{-1}{\bf z}_i](S_k+S_t)^{-1} \\
  &+\frac{n_k-1}{n_k}I({\bf z}_i\in C_k^B)(S_k+S_t)^{-1}(\overline{C}_k-\overline(C)_t){\bf z}_i^T(S_k+S_t)^{-1}]
\end{aligned}
\end{equation}

{\bf The final gradients.} { Using Eqs. \ref{eq:f06} and \ref{eq:f14}}, the final gradient of the objective for the developed statistical loss can be formulated as
\begin{equation}\label{eq:f15}
\begin{aligned}
  &\frac{\partial L}{{\partial \bf z}_i}=\frac{\partial L_0}{\partial {\bf z}_i}+\lambda\frac{\partial L_{div}}{\partial {\bf z}_i}=\frac{2}{\Lambda}\sum_{k=1}^{\Lambda}\frac{1}{n_k}I({\bf z}_i\in C_k^B)({\bf z}_i-\overline{C}_k)\\
  &+\frac{n_k+n_t-2}{\frac{1}{n_k}+\frac{1}{n_t}}[-\frac{2}{n_k}I({\bf z}_i\in C_k^B)(S_k+S_t)^{-1}(\overline{C}_k-\overline{C}_t)  \\
   &+\frac{n_k-1}{n_k}I({\bf z}_i\in C_k^B)[(\overline{C}_k-\overline{C}_t)^T(S_k+S_t)^{-1}{\bf z}_i](S_k+S_t)^{-1}  \\
   &+\frac{n_k-1}{n_k}I({\bf z}_i\in C_k^B)(S_k+S_t)^{-1}(\overline{C}_k-\overline{C}_t){\bf z}_i^T(S_k+S_t)^{-1}]
\end{aligned}
\end{equation}

\section{More Results}
All the additional experiments in the document use the same CNN architecture as that in the manuscript.

\subsection{Results on Salinas Scene Dataset}
\begin{figure*}[t]
\centering
 \subfigure[]{\includegraphics[width=0.2\linewidth]{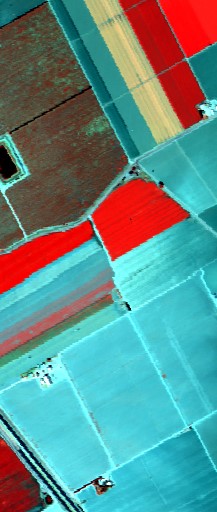}}
 \subfigure[]{\includegraphics[width=0.2\linewidth]{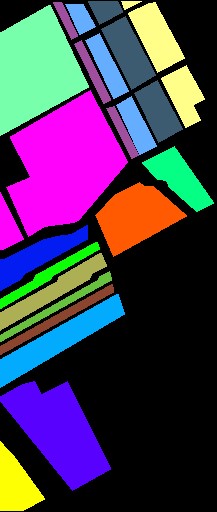}}
 \subfigure[]{\includegraphics[width=0.25\linewidth]{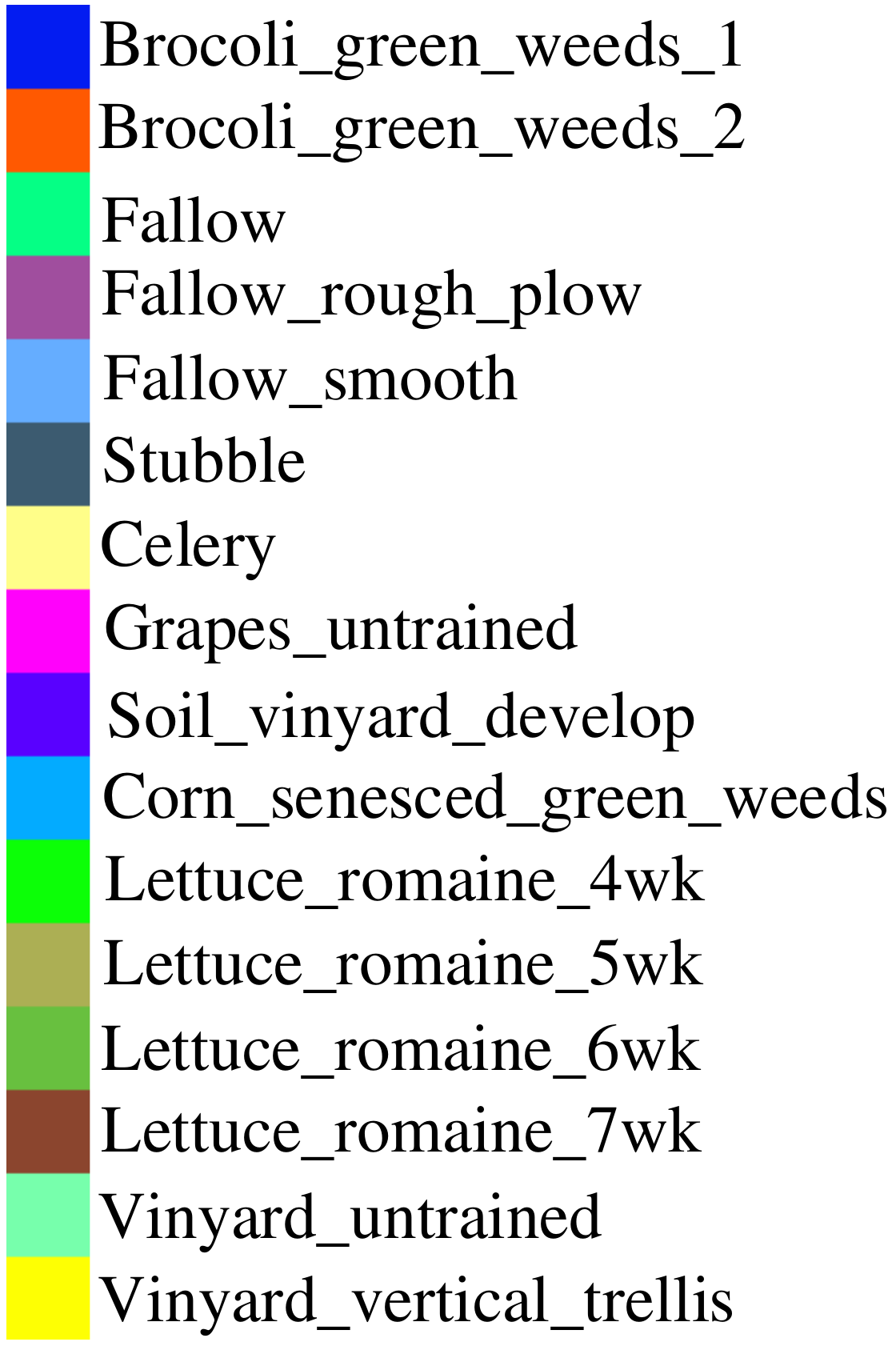}}
   \caption{Salinas scene dataset. (a) False color composite (band 40, 120, 180); (b) ground truth; (c) map color.}
\label{fig:salinas}
\end{figure*}

{\bf Experimental data set.} The Salinas Scene hyperspectral data set \cite{f01} collected over Salinas Valley in California was used to test the performance of the proposed method. The collected hyperspectral image has $217\times 512$ pixels with 224 bands and spatial resolution of 3.7m. 20 water absorption bands are abandoned and the remainder are used for experiments. A total of 54129 samples are selected from the image which can be divided into 16 classes. The false color composite and the ground truth can be seen in Fig. \ref{fig:salinas}. Since a large amount of samples are available in each class for experiments, if not specified, we select 200 samples per class for training and the remainder for testing.

{\bf General Performance.} { In this set of experiments, the diversity weight is set to 0.01. We set the iteration of the training process to 60000 and the learning rate is set to 0.001. Then, the proposed method took about 2182s over the Salinas Scene dataset}. Table \ref{table:salinas} shows the results of the SVM-POLY, the CNN with general softmax loss and the proposed statistical loss over the data. From the table, it can be noted that the CNN model by the proposed method can obtain an accuracy of 97.75\% $\pm$ 0.20\% which is remarkably improved than 97.01\% $\pm$ 0.22\% obtained by the CNN with general softmax loss. Moreover, compare the $|F_{ij}|$ in the McNemar's test, and we can also find that the $|F_{ij}|$ can achieve 11.23 which is much larger than 1.96. This means the improvement of the proposed method for the performance of hyperspectral image classification is statistically significant.

\begin{table}[t]
\begin{center}
\caption{Classification accuracies ($Mean\pm SD$) (OA, AA, and Kappa) of different methods achieved on the Salinas Scene data. The results from CNN is trained with the Softmax Loss. ${|F_{ij}|}$ represents the value of McNemar's test.}
\label{table:salinas}
\begin{tabular}{|c | c | c c c|}
%\toprule[1pt]
\hline
\multicolumn{2}{|c|}{\bf Methods}     &  {\bf SVM-POLY} &  {\bf CNN} &  {\bf Proposed Method} \\
\hline\hline
\multirow{16}{*}{\rotatebox{90}{\tabincell{c}{\textbf{Classification} \\ \textbf{Accuracies (\%)}}}}          & C1   &  $99.65\pm 0.15$  &  $99.89\pm 0.15$ &  $\mathbf{100.0\pm 0.00}$\\
                                                                                                             & C2   &  $99.87\pm 0.09$  &  $99.91\pm 0.09$ &  $\mathbf{100.0\pm 0.01}$\\
                                                                                                             & C3   &  $99.61\pm 0.17$  &  $99.98\pm 0.07$ &  $\mathbf{100.0\pm 0.00}$\\
                                                                                                             & C4   &  $99.53\pm 0.16$  &  $99.69\pm 0.27$ &  $\mathbf{99.91\pm 0.14}$\\
                                                                                                             & C5   &  $98.37\pm 0.51$  &  $99.53\pm 0.35$ &  $\mathbf{99.68\pm 0.19}$\\
                                                                                                             & C6   &  $99.77\pm 0.24$  &  $100.0\pm 0.00$ &  $\mathbf{100.0\pm 0.00}$\\
                                                                                                             & C7   &  $99.62\pm 0.23$  &  $99.86\pm 0.15$ &  $\mathbf{99.99\pm 0.02}$\\
                                                                                                             & C8   &  $79.13\pm 2.73$  &  $92.76\pm 1.85$ &  $\mathbf{95.23\pm 1.74}$\\
                                                                                                             & C9   &  $99.47\pm 0.39$  &  $99.90\pm 0.10$ &  $\mathbf{99.97\pm 0.05}$\\
                                                                                                             & C10   &  $93.25\pm 0.70$  &  $98.54\pm 0.66$ &  $\mathbf{99.05\pm 0.47}$\\
                                                                                                             & C11   &  $98.65\pm 0.80$  &  $99.38\pm 0.49$ &  $\mathbf{99.99\pm 0.04}$\\
                                                                                                             & C12   &  $99.93\pm 0.05$  &  $99.91\pm 0.15$ &  $\mathbf{100.0\pm 0.00}$\\
                                                                                                             & C13   &  $99.05\pm 0.45$  &  $99.92\pm 0.19$ &  $\mathbf{99.92\pm 0.07}$\\
                                                                                                             & C14   &  $97.06\pm 0.74$  &  $99.75\pm 0.37$ &  $\mathbf{99.82\pm 0.20}$\\
                                                                                                             & C15   &  $73.77\pm 1.96$  &  $91.10\pm 2.70$ &  $\mathbf{91.92\pm 1.83}$\\
                                                                                                             & C16   &  $99.09\pm 0.33$  &  $99.52\pm 0.48$ &  $\mathbf{99.79\pm 0.22}$\\
 \hline
 \multicolumn{2}{|c|}{{\bf OA}  (\%)}      &  $91.07\pm 0.42$ &  $97.01\pm 0.22$  &  $\mathbf{97.75\pm 0.20}$\\
 \hline
 \multicolumn{2}{|c|}{{\bf AA}  (\%)}      &  $95.99\pm 0.13$ &  $98.73\pm 0.12$  &  $\mathbf{99.08\pm 0.07}$\\
 \hline
 \multicolumn{2}{|c|}{{\bf KAPPA} (\%)}    &  $90.01\pm 0.46$ &  $96.65\pm 0.25$  &  $\mathbf{97.48\pm 0.22}$\\
\hline
\multicolumn{2}{|c|}{{${|F_{ij}|}$}}    &  $51.86$ &  $11.23$  &  $-$\\
\hline
%\bottomrule[1pt]
\end{tabular}
\end{center}
\end{table}

\begin{figure*}[t]
\centering
 \subfigure[]{\label{fig:2a}{\includegraphics[width=0.48\linewidth]{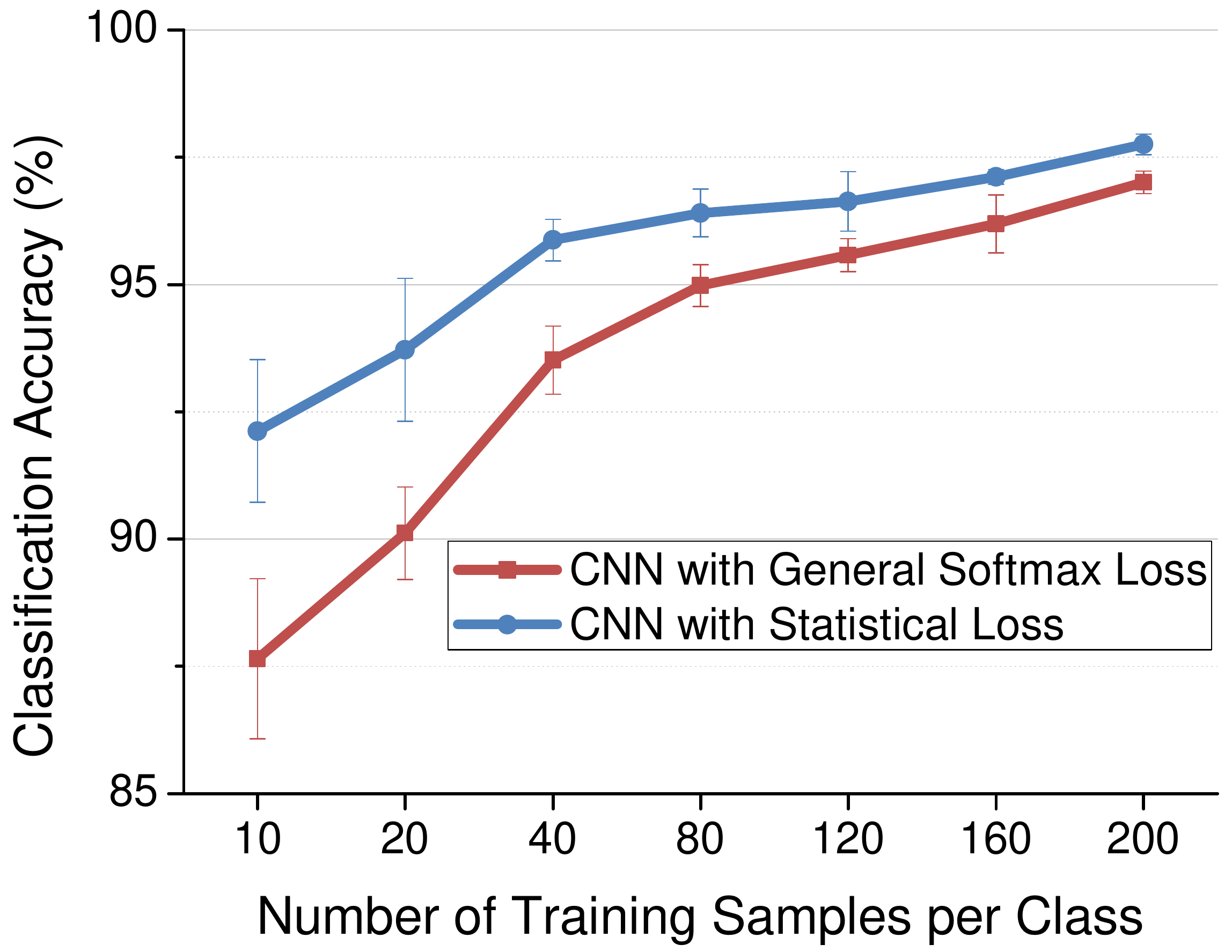}}}
 \subfigure[]{\label{fig:2b}{\includegraphics[width=0.48\linewidth]{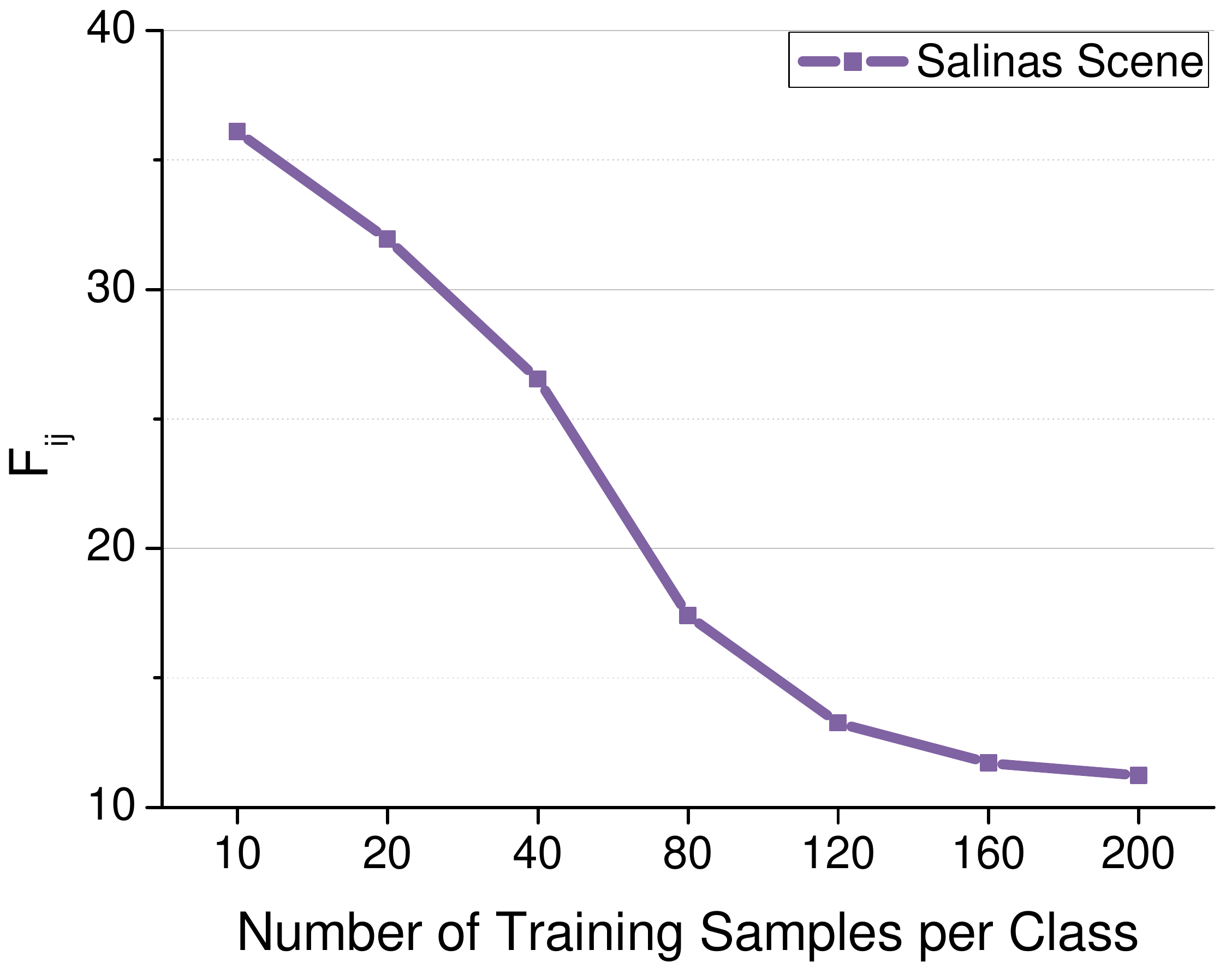}}}
   \caption{Classification performance of the proposed method over Salinas Scene data with different number of training samples. (a) Classification performance of the proposed method and CNN with general softmax loss; (b) McNemar's test between the proposed method and the general softmax loss.}
\label{fig:salinas_result}
\end{figure*}

{\bf Effects of Different Number of Training Samples.} In this set of experiments, the diversity weight is also set to 0.01. The number of training samples is selected from the set of $\{10,20,40,80,120,160,200\}$. Fig. \ref{fig:2a} shows the classification performance of the proposed method and the CNN with general softmax loss under different number of training samples. Fig. \ref{fig:2b} shows the corresponding value of McNemar's test  between the proposed method and the CNN with general softmax loss. We can find that the developed method can significantly improve the representational ability of the learned model for the hyperspectral image. When the number of training samples is set to 200, the classification error can be decreased by 24.75\%. Besides, the classification error can be decreased from 12.35\% to 7.88\% under 10 training samples per class. Moreover, from Fig. \ref{fig:2b}, we can also find that the value $|F_{ij}|$ is remarkably increased when reducing the number of training samples which demonstrates that the proposed method can also be effective for the task with limited training samples. Especially, when the number of training samples is set to 10, the value of $|F_{ij}|$ achieves 36.09.

\begin{figure*}[t]
\centering
 \subfigure[]{\label{fig:3a}{\includegraphics[width=0.153\linewidth]{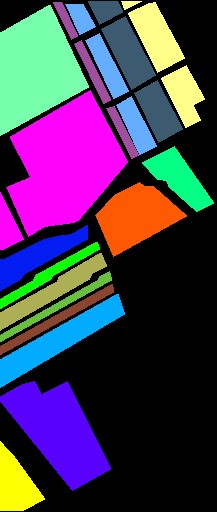}}}
 \subfigure[]{\label{fig:3b}{\includegraphics[width=0.153\linewidth]{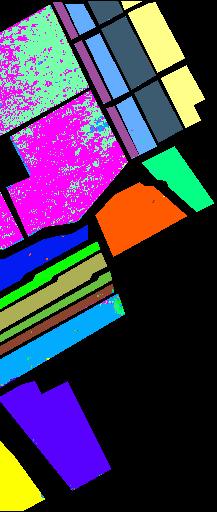}}}
 \subfigure[]{\label{fig:3c}{\includegraphics[width=0.153\linewidth]{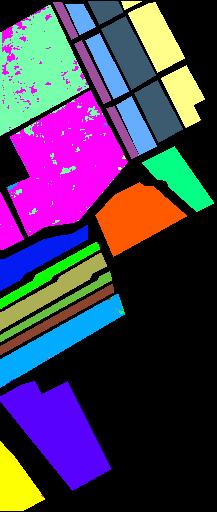}}}
 \subfigure[]{\label{fig:3d}{\includegraphics[width=0.153\linewidth]{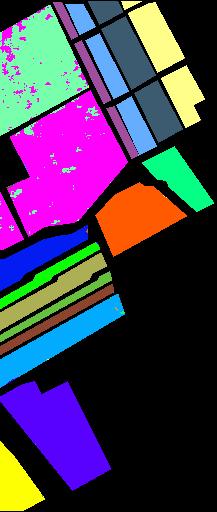}}}
 \subfigure[]{\label{fig:3e}{\includegraphics[width=0.153\linewidth]{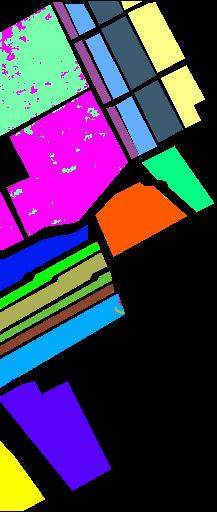}}}
 \subfigure[]{\label{fig:3f}{\includegraphics[width=0.153\linewidth]{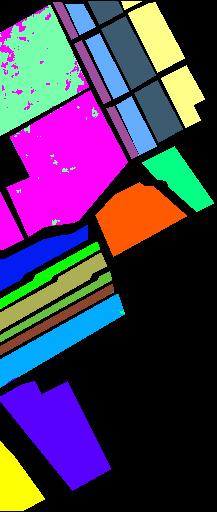}}}
 %\subfigure[]{\label{fig:3g}{\includegraphics[width=0.115\linewidth]{data/3g.jpg}}}
 %\subfigure[]{\label{fig:3h}{\includegraphics[width=0.115\linewidth]{data/3h.jpg}}}
   \caption{Salinas scene classification maps of different methods  with 200 samples per class for training (overall accuracies). (a) ground truth;  (b) SVM (91.16\%); (c) CNN with softmax loss (97.19\%); (d) CNN with center loss (97.54\%); (e) CNN with structured loss (97.53\%); (f) CNN with the developed statistical loss (97.97\%).}
\label{fig:salinas_map}
\end{figure*}

Furthermore, Fig. \ref{fig:salinas_map} presents the classification maps of the Salinas Scene data from different methods. Compare Fig. \ref{fig:3b} with \ref{fig:3f}, and \ref{fig:3c} with \ref{fig:3f}  and we can find that the proposed method can remarkably decrease the classification errors of the handcrafted method as well as the CNN with general softmax loss.

\begin{figure}[t]
\centering
   \includegraphics[width=0.9\linewidth]{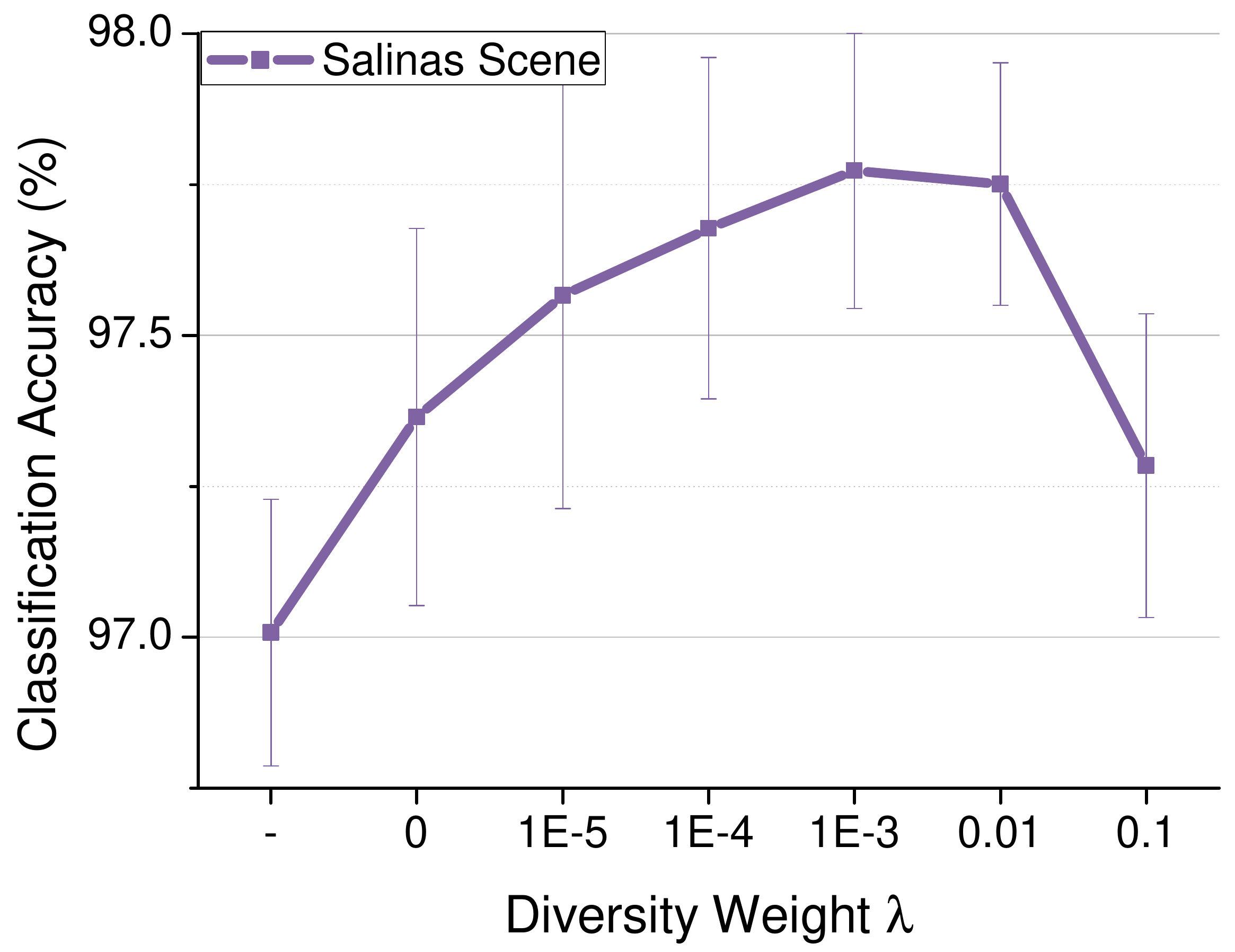}
   \caption{Classification performance of the proposed method with different diversity weight $\lambda$ over Salinas Scene data. '-' represents the results obtained with general softmax loss only.}
\label{fig:salinas_diversity}
\end{figure}

{\bf Effects of Diversity Weight $\lambda$}. The classification performance with different diversity weight $\lambda$ is shown in Fig. \ref{fig:salinas_diversity}. In the experiments, the value of $\lambda$ is chosen from $\{0,10^{-5},10^{-4},10^{-3},10^{-2},10^{-1}\}$. From the figure, we can find that the performance of the proposed method increases with the increase of $\lambda$. However, an extensively large $\lambda$ also negatively affects the performance of the proposed method. Especially, when $\lambda$ is set to 0.001, the proposed method can obtain an accuracy of 97.77\% $\pm$ 0.23\% which performs the best over the Salinas Scene data.

{\bf Comparisons with Other Samples-based Methods.} This work compares the performance of the developed statistical loss with the center loss \cite{31} and the structured loss \cite{30} which are selected as representative methods to model the correlation between the samples. Table \ref{table:salinas_comparisons_class_sample} shows the comparison results over the salinas data. It can be noted from the table that the developed method which can obtain an accuracy of $97.75\%\pm 0.20\%$ outperforms  the center loss ($97.43\%\pm 0.27\%$) and the structured loss ($97.40\%\pm 0.23\%$). Besides, the $|F_{ij}|$ achieves 5.35 when compare the CNN learned by the statistical loss with the CNN by the center loss and 5.62 when compare with that by the structured loss. This means that the improvement of the developed statistical loss over the center loss and the structured loss is statistically significant. Furthermore, Fig. \ref{fig:3d}, \ref{fig:3e} and \ref{fig:3f} shows the classification maps of the  CNN model by the center loss, the structured loss, and the statistical loss, respectively. It can be also noted that the developed statistical loss can better model the hyperspectral image than these samples-based methods.

\begin{table*}[t]
\begin{center}
\caption{Comparisons of the developed statistical loss with other samples-based methods on the Salinas Scene data. The center loss \cite{31} and the structured loss \cite{30} are chosen as baselines. It should be noted that the center loss and the structured loss are conducted with the joint learning of softmax loss.}
\vspace{1ex}
\label{table:salinas_comparisons_class_sample}
\begin{tabular}{|c | c | c c c c|}
%\toprule[1pt]
\hline
\multicolumn{2}{|c|}{\bf Methods}     &  {\bf Softmax Loss} &  {\bf Center Loss} &  {\bf Structured Loss} &  {\bf Proposed Method} \\
\hline\hline
\multirow{16}{*}{\rotatebox{90}{\tabincell{c}{\textbf{Classification} \\ \textbf{Accuracies (\%)}}}}          & C1   &    $99.89\pm 0.15$ & $99.94\pm 0.12$ & $99.96\pm 0.06$ &  $\mathbf{100.0\pm 0.00}$\\
                                                                                                             & C2   &    $99.91\pm 0.09$ & $99.95\pm 0.06$ & $99.96\pm 0.07$ &  $\mathbf{100.0\pm 0.01}$\\
                                                                                                             & C3   &    $99.98\pm 0.07$ & $100.0\pm 0.00$ & $100.0\pm 0.00$ &  $\mathbf{100.0\pm 0.00}$\\
                                                                                                             & C4   &    $99.69\pm 0.27$ & $99.74\pm 0.21$ & $99.81\pm 0.20$ &  $\mathbf{99.91\pm 0.14}$\\
                                                                                                             & C5   &    $99.53\pm 0.35$ & $\mathbf{99.70\pm 0.26}$ & $99.60\pm 0.30$ &  ${99.68\pm 0.19}$\\
                                                                                                             & C6   &    $100.0\pm 0.00$ & $100.0\pm 0.01$ & $100.0\pm 0.00$ &  $\mathbf{100.0\pm 0.00}$\\
                                                                                                             & C7   &    $99.86\pm 0.15$ & $99.89\pm 0.06$ & $99.92\pm 0.06$ &  $\mathbf{99.99\pm 0.02}$\\
                                                                                                             & C8   &    $92.76\pm 1.85$ & $93.97\pm 1.83$ & $93.58\pm 1.83$ &  $\mathbf{95.23\pm 1.74}$\\
                                                                                                             & C9   &    $99.90\pm 0.10$ & $99.84\pm 0.26$ & $99.85\pm 0.29$ &  $\mathbf{99.97\pm 0.05}$\\
                                                                                                             & C10   &    $98.54\pm 0.66$ & $\mathbf{99.24\pm 0.38}$ & $98.77\pm 0.59$ &  ${99.05\pm 0.47}$\\
                                                                                                             & C11   &    $99.38\pm 0.49$ & $99.76\pm 0.26$ & $99.62\pm 0.27$ &  $\mathbf{99.99\pm 0.04}$\\
                                                                                                             & C12   &    $99.91\pm 0.15$ & $99.98\pm 0.04$ & $99.99\pm 0.02$ &  $\mathbf{100.0\pm 0.00}$\\
                                                                                                             & C13   &    $99.92\pm 0.19$ & $99.96\pm 0.09$ & $\mathbf{99.96\pm 0.13}$ &  ${99.92\pm 0.07}$\\
                                                                                                             & C14   &    $99.75\pm 0.37$ & $\mathbf{99.89\pm 0.15}$ & $99.79\pm 0.32$ &  ${99.82\pm 0.20}$\\
                                                                                                             & C15   &    $91.10\pm 2.70$ & $91.77\pm 2.61$ & $\mathbf{92.41\pm 2.32}$ &  ${91.92\pm 1.83}$\\
                                                                                                             & C16   &    $99.52\pm 0.48$ & $99.55\pm 0.49$ & $99.64\pm 0.38$ &  $\mathbf{99.79\pm 0.22}$\\
 \hline
 \multicolumn{2}{|c|}{{\bf OA}  (\%)}      &    $97.01\pm 0.22$  & $97.43\pm 0.27$ & $97.40\pm 0.23$ &  $\mathbf{97.75\pm 0.20}$\\
 \hline
 \multicolumn{2}{|c|}{{\bf AA}  (\%)}      &    $98.73\pm 0.12$  & $98.95\pm 0.12$ & $98.93\pm 0.10$ &  $\mathbf{99.08\pm 0.07}$\\
 \hline
 \multicolumn{2}{|c|}{{\bf KAPPA} (\%)}    &    $96.65\pm 0.25$  & $97.12\pm 0.30$ & $97.09\pm 0.25$ &  $\mathbf{97.48\pm 0.22}$\\
\hline
\multicolumn{2}{|c|}{{${|F_{ij}|}$}}    &   $11.23$  & $5.35$  & $5.62$  &  $-$\\
\hline
%\bottomrule[1pt]
\end{tabular}
\end{center}
\end{table*}

{\bf Comparisons with the Most Recent Methods.} For Salinas Scene data, the results from the developed method are compared with that from the most recent methods: CNN-PPF \cite{15}, Contextual DCNN \cite{04}, DPP-DML-MS-CNN \cite{01}, and Spec-Spat \cite{f06}. Table \ref{table:salinas_comparison} lists the comparison results from different methods. All the results in the table are with the same experimental setups. We can find that the proposed method which can obtain an accuracy of 97.75\% $\pm$ 0.20\% outperforms these state-of-the-art methods. Moreover, we choose the NFE method in \cite{49} which focus on the task of limited training samples as baseline to validate the effectiveness of the proposed method with limited training samples. We can find the NFE can obtain 87.69\% OA, 93.93\% AA, and 86\% KAPPA with 15 training samples per class while the proposed method can obtain 92.12\% OA, 96.27\% AA and 91.24\% KAPPA with only 10 training samples per class. This indicates that the proposed method can also be applied for the training of the CNN model with limited training samples.

\begin{table*}[t]
\begin{center}
\caption{Classification performance of different methods with spectral-spatial information of Salinas Scene data in the most recent literature(200 training samples per class for training). }
\label{table:salinas_comparison}
\begin{tabular}{| c | c | c | c |}
%\toprule[1pt]
\hline
{\bf Methods}     &  {\bf OA(\%)} &  {\bf AA(\%)} &  {\bf KAPPA(\%)} \\
\hline\hline

{\bf SVM-POLY} &  $91.07\pm 0.42$  &  $95.99\pm 0.13$ &  ${90.01\pm 0.46}$\\
{\bf CNN-PPF \cite{15}}    &  $94.80$ &  $-$  &  $-$\\
 {\bf Contextual DCNN \cite{04}} &  $95.07\pm 0.23$ &  $-$  &  $-$\\
  {\bf Spec-Spat \cite{f06}}    &  $96.07$ &  ${97.56}$  &  ${96.78}$\\
{\bf DPP-DML-MS-CNN \cite{01}}    &  $97.51\pm 0.18$ &  ${98.85\pm 0.05}$  &  ${97.88\pm 0.23}$\\
  {\bf Proposed Method} &  $\mathbf{97.75\pm 0.20}$ &  $\mathbf{99.08\pm 0.07}$  &  $\mathbf{97.48\pm 0.22}$\\
\hline

%\bottomrule[1pt]
\end{tabular}
\end{center}
\end{table*}

\subsection{Results on Kennedy Space Center (KSC)}

\begin{figure*}[t]
\centering
 \subfigure[]{\includegraphics[width=0.35\linewidth]{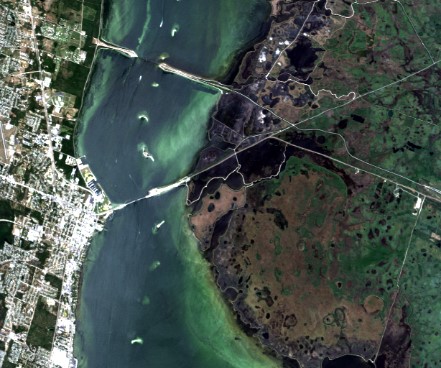}}
 \subfigure[]{\includegraphics[width=0.35\linewidth]{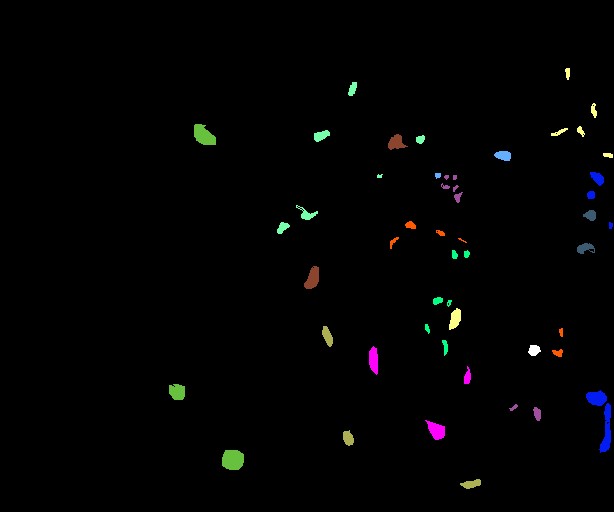}}
 \subfigure[]{\includegraphics[width=0.12\linewidth]{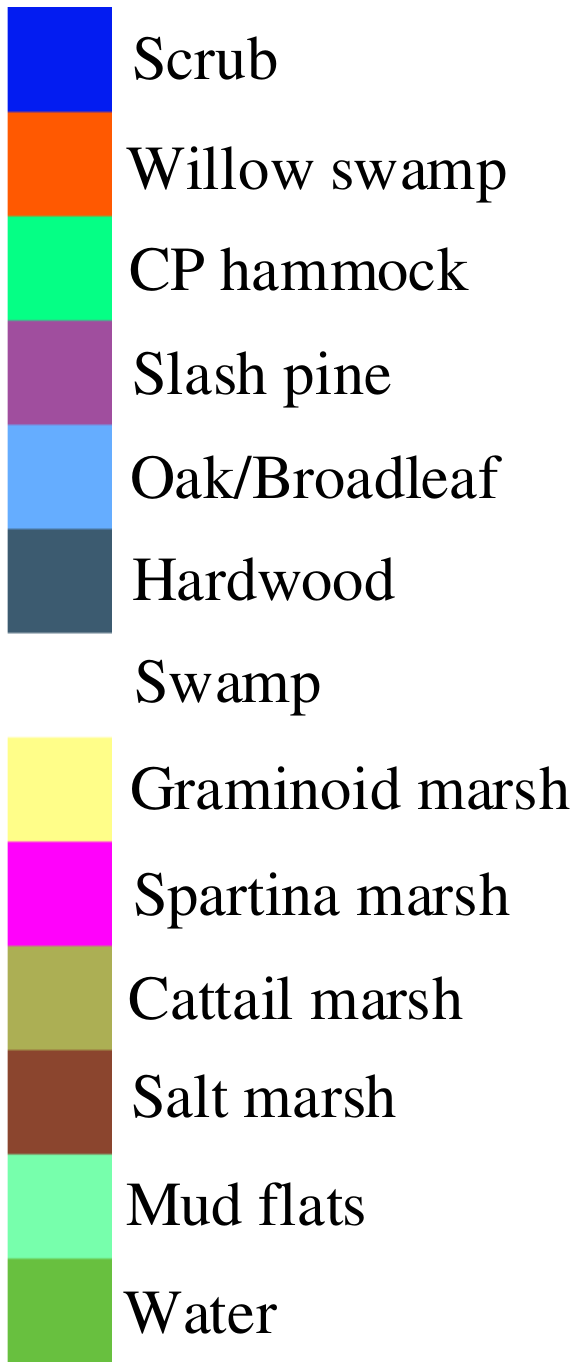}}
   \caption{Kennedy Space Center (KSC) dataset. (a) False color composite (band 28, 19, 10); (b) ground truth; (c) map color.}
\label{fig:ksc}
\end{figure*}

{\bf Experimental Data Set.} The KSC hyperspectral image \cite{f02} was obtained with the NASA Airborne Visible/Infrared Imaging Spectrometer (AVIRIS) over the Kennedy Space Center (KSC), Florida, on March 23, 1996. It consists of $614\times 512$ pixels which have a resolution of 18m with 224 bands ranging from 0.4 to 2.5 $\mu m$. Due to the water absorption and low SNR, 176 bands are remained for the analysis. For classification purposes, 5211 labeled samples divided into 13 classes are selected for experiments in this paper. Fig. \ref{fig:ksc} shows the false color composite and the ground truth of the dataset.

\begin{table}[t]
\begin{center}
\caption{Classification accuracies ($Mean\pm SD$) (OA, AA, and Kappa) of different methods achieved on the KSC data (with 10\% of samples for training). }
\label{table:ksc}
\begin{tabular}{|c | c | c c c|}
%\toprule[1pt]
\hline
\multicolumn{2}{|c|}{\bf Methods}     &  {\bf SVM-POLY} &  {\bf CNN} &  {\bf Proposed Method} \\
\hline\hline
\multirow{13}{*}{\rotatebox{90}{\tabincell{c}{\textbf{Classification} \\ \textbf{Accuracies (\%)}}}}          & C1   &  $89.75\pm 2.65$  &  $97.46\pm 0.89$ &  $\mathbf{98.80\pm 0.82}$\\
                                                                                                             & C2   &  $86.14\pm 2.83$  &  $95.00\pm 2.74$ &  $\mathbf{97.86\pm 3.23}$\\
                                                                                                             & C3   &  $90.60\pm 3.59$  &  $94.01\pm 1.93$ &  $\mathbf{96.98\pm 2.84}$\\
                                                                                                             & C4   &  $64.96\pm 4.78$  &  $81.36\pm 8.80$ &  $\mathbf{93.55\pm 3.02}$\\
                                                                                                             & C5   &  $61.16\pm 9.05$  &  $81.44\pm 6.75$ &  $\mathbf{85.96\pm 6.39}$\\
                                                                                                             & C6   &  $62.66\pm 5.04$  &  $91.84\pm 4.83$ &  $\mathbf{95.22\pm 4.52}$\\
                                                                                                             & C7   &  $85.21\pm 4.83$  &  $\mathbf{99.06\pm 1.04}$ &  ${97.08\pm 4.65}$\\
                                                                                                             & C8   &  $86.36\pm 3.25$  &  $98.75\pm 1.06$ &  $\mathbf{99.36\pm 0.69}$\\
                                                                                                             & C9   &  $94.16\pm 3.00$  &  $99.91\pm 0.21$ &  $\mathbf{100.0\pm 0.00}$\\
                                                                                                             & C10   &  $90.08\pm 3.20$  &  $99.37\pm 1.90$ &  $\mathbf{100.0\pm 0.00}$\\
                                                                                                             & C11   &  $95.34\pm 1.58$  &  $98.73\pm 2.75$ &  $\mathbf{99.95\pm 0.11}$\\
                                                                                                             & C12   &  $91.78\pm 2.04$  &  $99.53\pm 0.73$ &  $\mathbf{99.89\pm 0.24}$\\
                                                                                                             & C13   &  $99.87\pm 0.19$  &  $100.0\pm 0.00$ &  $\mathbf{100.0\pm 0.00}$\\
 \hline
 \multicolumn{2}{|c|}{{\bf OA}  (\%)}      &  $88.89\pm 0.49$ &  $96.94\pm 0.66$  &  $\mathbf{98.49\pm 0.45}$\\
 \hline
 \multicolumn{2}{|c|}{{\bf AA}  (\%)}      &  $84.47\pm 0.54$ &  $95.11\pm 1.21$  &  $\mathbf{97.28\pm 0.88}$\\
 \hline
 \multicolumn{2}{|c|}{{\bf KAPPA} (\%)}    &  $87.62\pm 0.54$ &  $96.59\pm 0.73$  &  $\mathbf{98.32\pm 0.51}$\\
\hline
\multicolumn{2}{|c|}{{${|F_{ij}|}$}}    &  $20.50$ &  $6.69$  &  $-$\\
\hline
%\bottomrule[1pt]
\end{tabular}
\end{center}
\end{table}

\begin{figure*}[t]
\centering
 \subfigure[]{\label{fig:6a}{\includegraphics[width=0.48\linewidth]{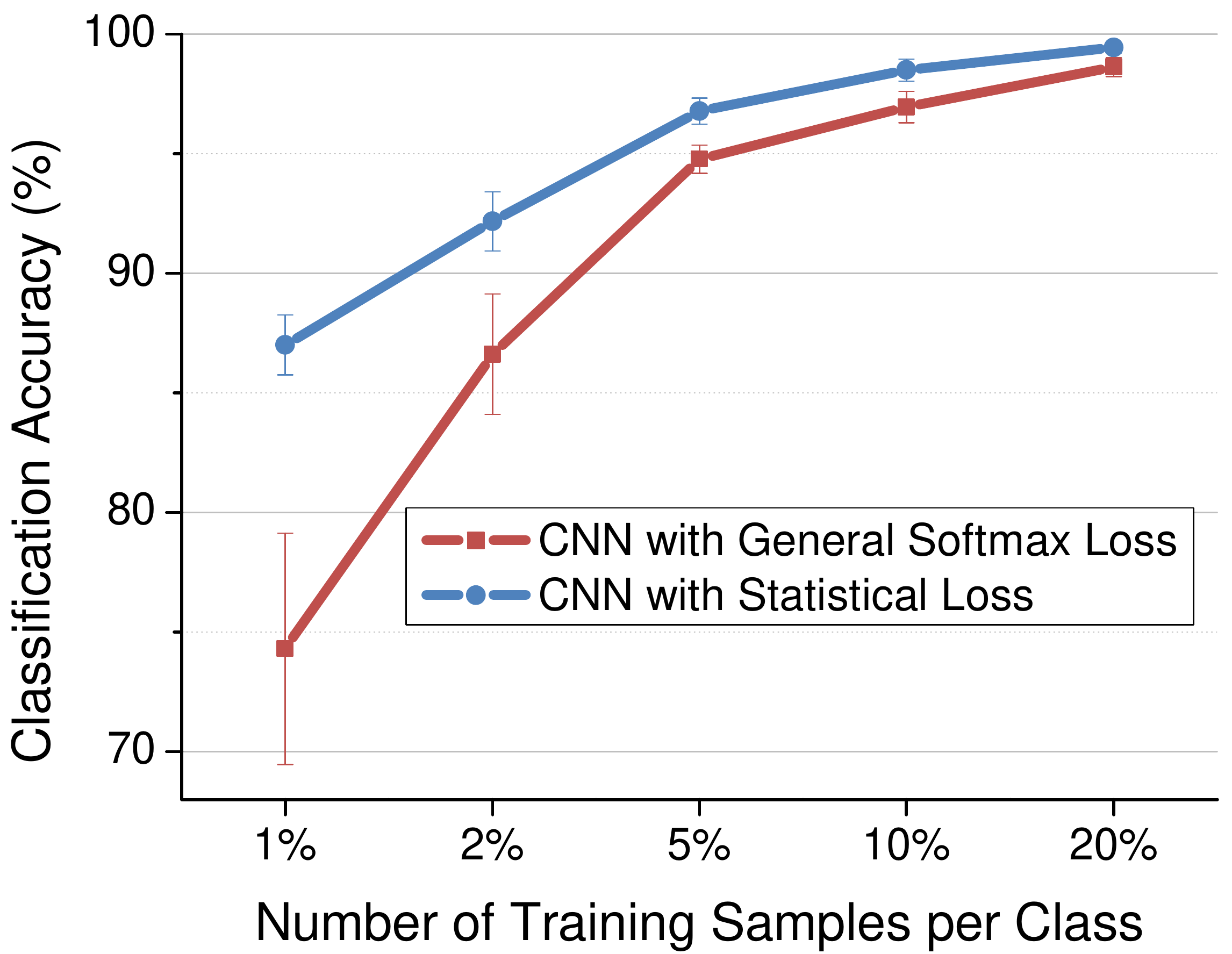}}}
 \subfigure[]{\label{fig:6b}{\includegraphics[width=0.48\linewidth]{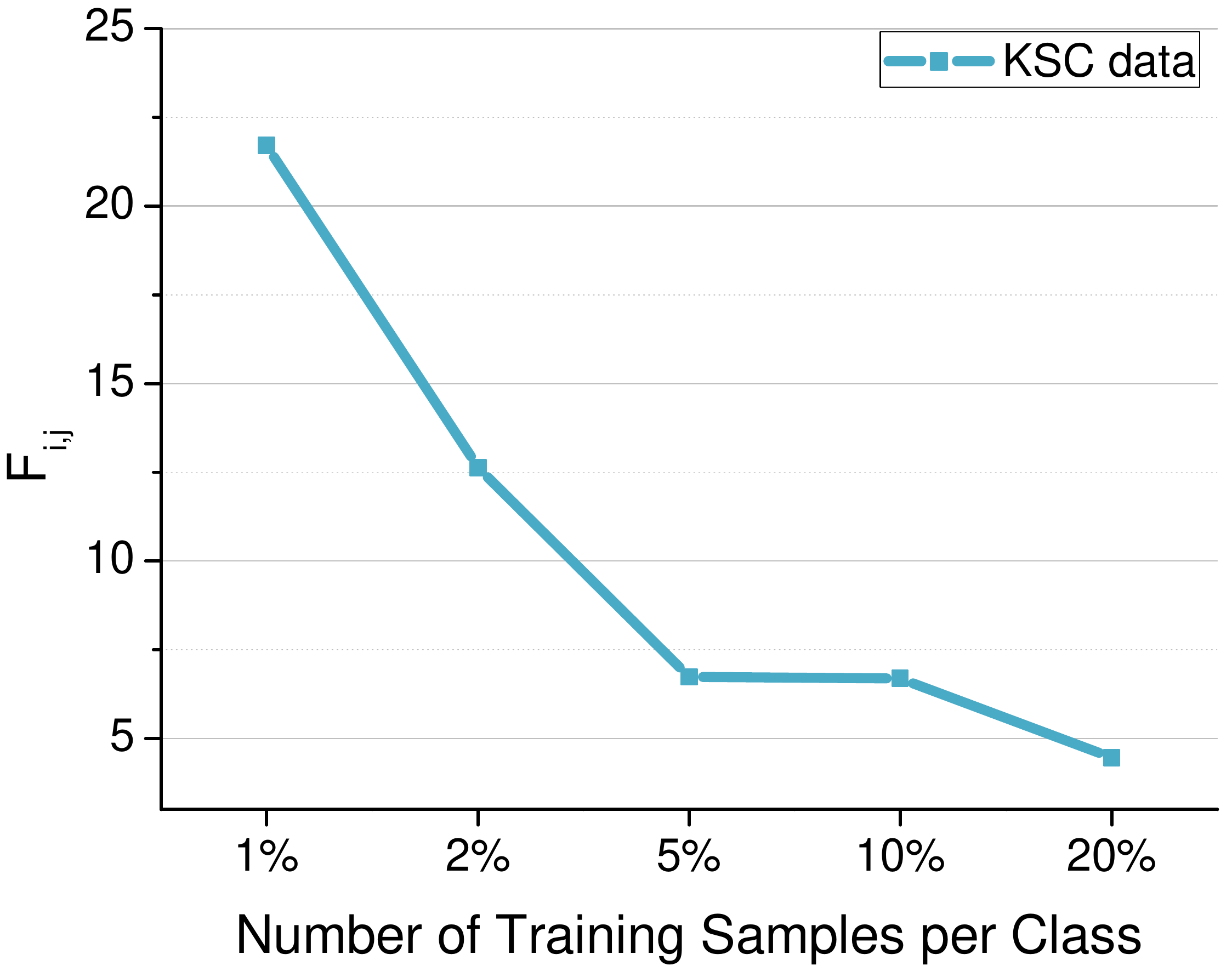}}}
   \caption{Classification performance of the proposed method over KSC data with different number of training samples. (a) Classification performance of the proposed method and CNN with general softmax loss; (b) McNemar's test between the proposed method and the general softmax loss.}
\label{fig:ksc_result}
\end{figure*}

\begin{figure*}[t]
\centering
 \subfigure[]{\label{fig:7a}{\includegraphics[width=0.3\linewidth]{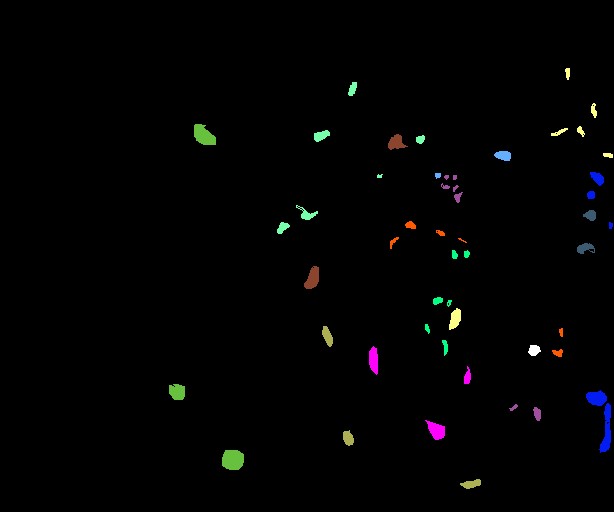}}}
 \subfigure[]{\label{fig:7b}{\includegraphics[width=0.3\linewidth]{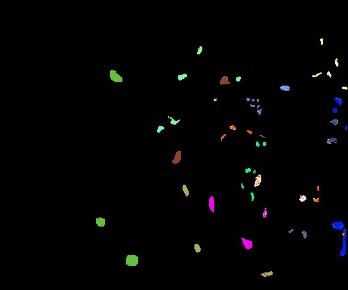}}}
 \subfigure[]{\label{fig:7c}{\includegraphics[width=0.3\linewidth]{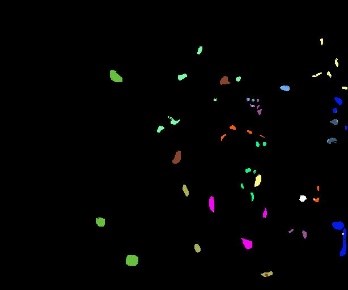}}}
 \subfigure[]{\label{fig:7d}{\includegraphics[width=0.3\linewidth]{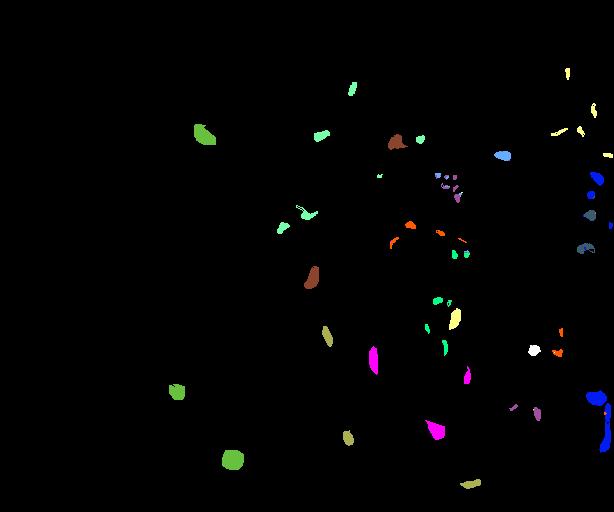}}}
 \subfigure[]{\label{fig:7e}{\includegraphics[width=0.3\linewidth]{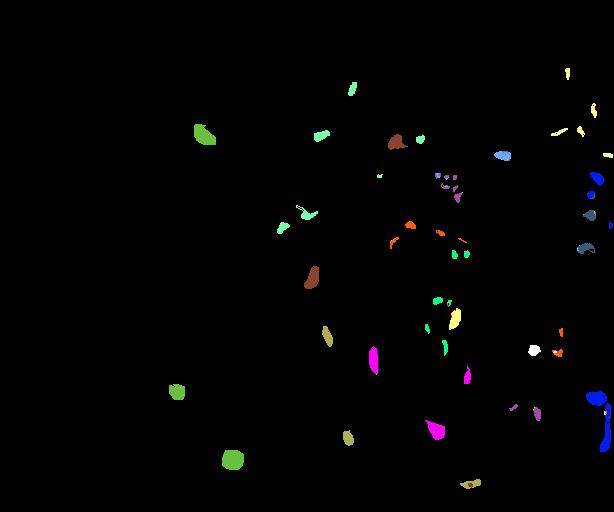}}}
 \subfigure[]{\label{fig:7f}{\includegraphics[width=0.3\linewidth]{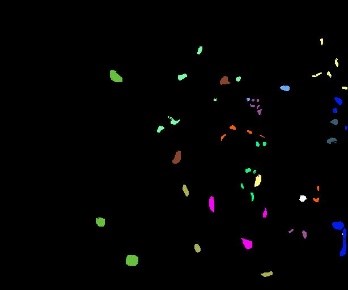}}}
 %\subfigure[]{\label{fig:7g}{\includegraphics[width=0.236\linewidth]{data/7g.jpg}}}
 %\subfigure[]{\label{fig:7h}{\includegraphics[width=0.236\linewidth]{data/7h.jpg}}}
 %\subfigure[]{\label{fig:7i}{\includegraphics[width=0.236\linewidth]{data/7i.jpg}}}
 %\subfigure[]{\label{fig:7j}{\includegraphics[width=0.236\linewidth]{data/7j.jpg}}}
 %\subfigure[]{\label{fig:7k}{\includegraphics[width=0.236\linewidth]{data/7k.jpg}}}
 %\subfigure[]{\label{fig:7l}{\includegraphics[width=0.236\linewidth]{data/7l.jpg}}}
   \caption{KSC classification maps of different methods with 10\% training samples per class (overall accuracies). (a) ground truth; (b) SVM (89.44\%); (c) CNN with softmax loss (96.86\%); (d) CNN with center loss (98.30\%); (e) CNN with structured loss (97.14\%); (f) proposed method (99.01\%).}
\label{fig:ksc_map}
\end{figure*}

\begin{figure}[t]
\centering
   \includegraphics[width=0.9\linewidth]{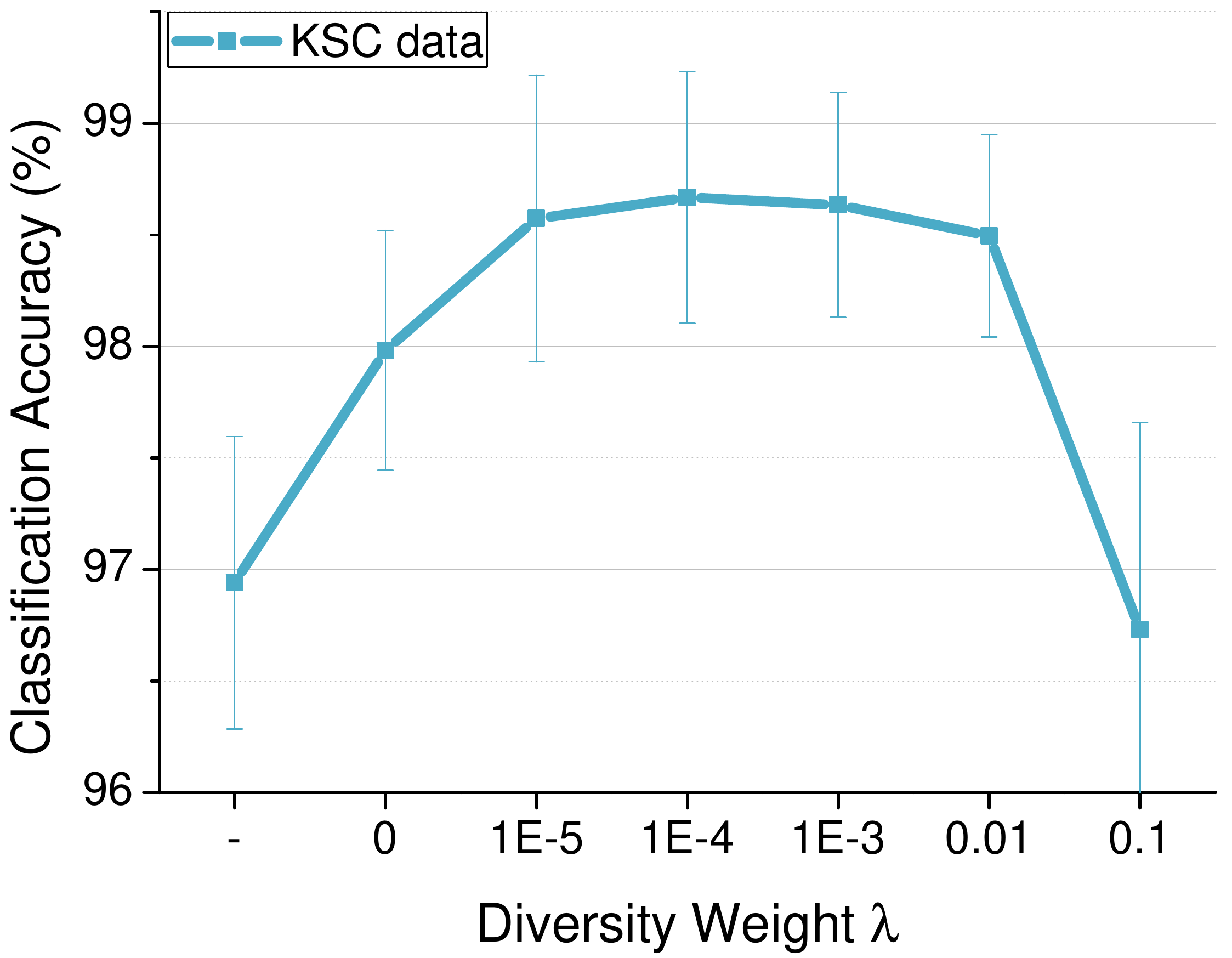}
   \caption{Classification performance of the proposed method with different diversity weight $\lambda$ over KSC data. '-' represents the results obtained with general softmax loss only.}
\label{fig:ksc_diversity}
\end{figure}

{\bf The Experimental Results.} First, we set the diversity weight, the iteration of the training process, and the learning rate to 0.01, 60000, and 0.001, respectively. We train the CNN model with 10\% of samples and the remainder for test. { We present the general performance of the proposed method over the KSC data. The proposed method took about 1789s over the KSC data.} Table \ref{table:ksc} shows the classification accuracy of each class of the SVM, the CNN with general softmax loss as well as the proposed method. The developed method can make statistically significant improvement over the learned model for hyperspectral image. Besides, Fig. \ref{fig:6a} shows the classification performance of the proposed method and the CNN with general softmax loss and Fig. \ref{fig:6b} presents the corresponding McNemar's test. In this set of experiments, the diversity weight $\lambda$ is set to 0.01. From Fig. \ref{fig:ksc_result}, we can find that it presents the similar tendencies as Salinas Scene data. Especially, the proposed method can improve the classification accuracy from 74.30\% to 87.00\% under $1\%$ of samples for training over the KSC data. This also demonstrates the effectiveness of the proposed method over the task with limited training samples.
Furthermore, Fig. \ref{fig:ksc_map} presents the classification maps of different methods over the KSC data. The comparisons of the maps from different methods in Fig. \ref{fig:ksc_map} further indicate the effectiveness of the proposed method.
Besides, Fig. \ref{fig:ksc_diversity} shows the classification performance of the proposed method with different diversity weight $\lambda$ over the KSC data. The performance of the proposed method also increases with the increase of the diversity weight. Similar to the tendencies on other datasets, an extensively large $\lambda$ shows negative effects on the performance of the proposed method. It should be noted from the Fig. \ref{fig:ksc_diversity} that when $\lambda=0.0001$, the proposed method performs the best which can achieve 98.67\% $\pm$ 0.56\% OA over the KSC data.

{\bf Comparisons with Other Samples-based Methods.}
We also compare the developed statistical loss with other recent methods which model the feature correlation between the samples. The recently developed center loss \cite{31} and structured loss \cite{30} are chosen as the baselines. The comparison results are shown in Table \ref{table:ksc_comparisons_class_sample}. From the table, we can find that the developed statistical loss which is formulated with the class distributions outperforms the recent samples-based losses, such as the center loss and the structured loss.
%It should also be noted that the lifted loss even cannot be fit for the KSC data and cannot improve the performance of the learned model for the KSC data.

\begin{table*}[t]
\begin{center}
\caption{Comparisons of the developed statistical loss with other samples-based methods on the KSC data. The center loss \cite{31} and the structured loss \cite{30} are chosen as baselines. }
\label{table:ksc_comparisons_class_sample}
\begin{tabular}{|c | c | c c c c|}
%\toprule[1pt]
\hline
\multicolumn{2}{|c|}{\bf Methods}     &  {\bf Softmax Loss} &  {\bf Center Loss} &  {\bf Structured Loss} &  {\bf Statistical Loss} \\
\hline\hline
\multirow{13}{*}{\rotatebox{90}{\tabincell{c}{\textbf{Classification} \\ \textbf{Accuracies (\%)}}}}          & C1     &  $97.46\pm 0.89$ &  $98.10\pm 0.76$ &  $97.23\pm 1.32$ &  $\mathbf{98.80\pm 0.82}$\\
                                                                                                             & C2     &  $95.00\pm 2.74$ &  $96.59\pm 3.48$ &  $95.15\pm 2.95$ &  $\mathbf{97.86\pm 3.23}$\\
                                                                                                             & C3     &  $94.01\pm 1.93$ &  $\mathbf{98.02\pm 1.67}$ &  $94.40\pm 2.40$ &  ${96.98\pm 2.84}$\\
                                                                                                             & C4     &  $81.36\pm 8.80$ &  $90.18\pm 7.00$ &  $81.82\pm 7.71$ &  $\mathbf{93.55\pm 3.02}$\\
                                                                                                             & C5     &  $81.44\pm 6.75$ &  $84.66\pm 6.49$ &  $80.97\pm 8.40$ &  $\mathbf{85.96\pm 6.39}$\\
                                                                                                             & C6     &  $91.84\pm 4.83$ &  $93.09\pm 6.98$ &  $91.30\pm 7.25$ &  $\mathbf{95.22\pm 4.52}$\\
                                                                                                             & C7     &  $99.06\pm 1.04$ &  $97.60\pm 3.26$ &  $\mathbf{97.80\pm 3.43}$ &  ${97.08\pm 4.65}$\\
                                                                                                             & C8     &  $98.75\pm 1.06$ &  $\mathbf{99.39\pm 0.75}$ &  $98.87\pm 0.88$ &  ${99.36\pm 0.69}$\\
                                                                                                             & C9     &  $99.91\pm 0.21$ &  $99.98\pm 0.07$ &  $99.98\pm 0.07$ &  $\mathbf{100.0\pm 0.00}$\\
                                                                                                             & C10     &  $99.37\pm 1.90$ &  $99.95\pm 0.17$ &  $99.36\pm 1.92$ &  $\mathbf{100.0\pm 0.00}$\\
                                                                                                             & C11     &  $98.73\pm 2.75$ &  $99.34\pm 1.22$ &  $98.68\pm 2.34$ &  $\mathbf{99.95\pm 0.11}$\\
                                                                                                             & C12     &  $99.53\pm 0.73$ &  $99.47\pm 0.84$ &  $99.19\pm 1.45$ &  $\mathbf{99.89\pm 0.24}$\\
                                                                                                             & C13     &  $100.0\pm 0.00$ &  $100.0\pm 0.00$ &  $100.0\pm 0.00$ &  $\mathbf{100.0\pm 0.00}$\\
 \hline
 \multicolumn{2}{|c|}{{\bf OA}  (\%)}      &    $96.94\pm 0.66$ &  $98.00\pm 0.51$ &  $96.87\pm 0.75$  &  $\mathbf{98.49\pm 0.45}$\\
 \hline
 \multicolumn{2}{|c|}{{\bf AA}  (\%)}      &    $95.11\pm 1.21$ &  $96.64\pm 0.93$ &  $94.98\pm 1.44$  &  $\mathbf{97.28\pm 0.88}$\\
 \hline
 \multicolumn{2}{|c|}{{\bf KAPPA} (\%)}    &    $96.59\pm 0.73$ &  $97.77\pm 0.57$ &  $96.51\pm 0.83$  &  $\mathbf{98.32\pm 0.51}$\\
\hline
\multicolumn{2}{|c|}{{${|F_{ij}|}$}}    &   $6.69$  & $2.59$  & $7.23$  &  $-$\\
\hline
%\bottomrule[1pt]
\end{tabular}
\end{center}
\end{table*}

{\bf Comparisons with the Most Recent Methods.} For KSC data, we compare the proposed method with other state-of-the-art methods under 10\%, 20\% training samples, respectively. The comparison results are listed in Table \ref{table:ksc_comparison}. When using 10\% training samples, the proposed method can obtain 98.49\% $\pm$ 0.45\% OA (Actually, the proposed method obtains 98.67\% $\pm$ 0.56\% OA when $\lambda=0.0001$) which is better than the most recent methods, such as DPP-DML-MS-CNN (97.51\% $\pm$ 0.18\% OA) \cite{01}, FDA-SVM (93.68\% OA) \cite{f07}. Besides, when using 20\% training samples, the proposed method can obtain 99.43\% $\pm$ 0.21\% OA which is comparable or better than SSRN ($5\times 5$) (96.99\% $\pm$ 0.55\% OA) \cite{03} and DPP-DML-MS-CNN (99.42\% $\pm$ 0.18\% OA) \cite{01}. It should be noted that the proposed method can also obtain a comparable performance with only $5\times 5$ neighborhoods than SSRN with $11\times 11$ neighborhoods (99.61\% $\pm$ 0.22\% OA). The SSRN with $11\times 11$ neighborhoods requires more computational source (2466s) than the proposed method.

\begin{table*}[t]
\begin{center}
\caption{Classification performance of different methods with spectral-spatial information of KSC data in the most recent literature. The methods upwards the double line present the results with 10\% samples for training and the results under the double line present the results with 20\% samples for training.}
\label{table:ksc_comparison}
\begin{tabular}{| c | c | c | c |}
%\toprule[1pt]
\hline
{\bf Methods}     &  {\bf OA(\%)} &  {\bf AA(\%)} &  {\bf KAPPA(\%)} \\
\hline\hline

{\bf SVM-POLY} &  $88.89\pm 0.49$  &  $84.47\pm 0.54$ &  ${87.62\pm 0.54}$\\
{\bf FDA-SVM \cite{f07}}    &  $93.68$ &  $-$  &  $-$\\
 {\bf DPP-DML-MS-CNN \cite{01}} &  $97.51\pm 0.18$ &  $98.85\pm 0.05$  &  $97.88\pm 0.23$\\
  {\bf Proposed Method} &  $\mathbf{98.49\pm 0.45}$ &  $\mathbf{97.28\pm 0.88}$  &  $\mathbf{98.32\pm 0.51}$\\
  \hline\hline
{\bf DPP-DML-MS-CNN \cite{01}}    &  $99.42\pm 0.18$ &  ${98.91\pm 0.41}$  &  ${99.32\pm 0.23}$\\
{\bf SSRN ($5\times 5$) \cite{03}}    &  $96.99\pm 0.55$ &  ${-}$  &  ${-}$\\
{\bf SSRN ($11\times 11$) \cite{03}}    &  $\mathbf{99.61\pm 0.22}$ &  $\mathbf{99.33\pm 0.57}$  &  $\mathbf{99.56\pm 0.25}$\\
  {\bf Proposed Method} &  $\mathbf{99.43\pm 0.21}$ &  $\mathbf{98.80\pm 0.55}$  &  $\mathbf{99.37\pm 0.24}$\\
\hline

%\bottomrule[1pt]
\end{tabular}
\end{center}
\end{table*}

% or


%\appendices

%\section{}

%% use section* for acknowledgment
%\section*{Acknowledgment}
%
%This work was supported in part by the Natural Science
%Foundation of China under Grant 61671456 and 61271439, in
%part by the Foundation for the Author of National Excellent
%Doctoral Dissertation of China (FANEDD) under Grant
%201243, and in part by the Program for New Century Excellent
%Talents in University under Grant NECT-13-0164.

\ifCLASSOPTIONcaptionsoff
  \newpage
\fi



%\bibliographystyle{IEEEtran}

\bibliographystyle{ieee}
\bibliography{egbib}

\vfill


\end{document}